%% file: Formatting-Instructions-LaTeX-2024.tex
\documentclass[letterpaper]{article} % DO NOT CHANGE THIS
\usepackage{aaai24}  % DO NOT CHANGE THIS
\usepackage{times}  % DO NOT CHANGE THIS
\usepackage{helvet}  % DO NOT CHANGE THIS
\usepackage{courier}  % DO NOT CHANGE THIS
\usepackage[hyphens]{url}  % DO NOT CHANGE THIS
\usepackage{graphicx} % DO NOT CHANGE THIS
\usepackage{natbib}  % DO NOT CHANGE THIS AND DO NOT ADD ANY OPTIONS TO IT
\usepackage{caption} % DO NOT CHANGE THIS AND DO NOT ADD ANY OPTIONS TO IT
\usepackage{algorithm}
\usepackage{algorithmic}

\usepackage{newfloat}
\usepackage{listings}
\DeclareCaptionStyle{ruled}{labelfont=normalfont,labelsep=colon,strut=off} % DO NOT CHANGE THIS
\floatstyle{ruled}
\newfloat{listing}{tb}{lst}{}
\floatname{listing}{Listing}
\usepackage{microtype}
\usepackage{booktabs}       % professional-quality tables
\usepackage{multirow}
\usepackage{amsfonts}
\usepackage{amsmath}
\title{
Beyond Expected Return: Accounting for Policy Reproducibility When Evaluating Reinforcement Learning Algorithms
}
\author {
    Manon Flageat\footnote{Equal Contribution},
    Bryan Lim\footnotemark[\value{footnote}],
    Antoine Cully
}
\affiliations {
    Adaptive and Intelligent Robotics Lab, Imperial College London, UK\\
    manon.flageat18@imperial.ac.uk, bryan.lim16@imperial.ac.uk, a.cully@imperial.ac.uk
}

\usepackage{bibentry}
\begin{document}

\maketitle

\begin{abstract}
\input{abstract}
\end{abstract}

\input{main}

\newpage

\bibliography{aaai24}

\cleardoublepage
\appendix
\input{appendix}

\end{document}

%% file: abstract.tex
Many applications in Reinforcement Learning (RL) usually have noise or stochasticity present in the environment. 
Beyond their impact on learning, these uncertainties lead the exact same policy to perform differently, i.e. yield different return, from one roll-out to another. 
Common evaluation procedures in RL summarise the consequent return distributions using solely the expected return, which does not account for the spread of the distribution. 
Our work defines this spread as the \textit{policy reproducibility}: the ability of a policy to obtain similar performance when rolled out many times, a crucial property in some real-world applications. 
We highlight that existing procedures that only use the expected return are limited on two fronts: first an infinite number of return distributions with a wide range of performance-reproducibility trade-offs can have the same expected return, limiting its effectiveness when used for comparing policies; second, the expected return metric does not leave any room for practitioners to choose the best trade-off value for considered applications. 
In this work, we address these limitations by recommending the use of Lower Confidence Bound, a metric taken from Bayesian optimisation that provides the user with a preference parameter to choose a desired performance-reproducibility trade-off.
We also formalise and quantify \textit{policy reproducibility}, and demonstrate the benefit of our metrics using extensive experiments of popular RL algorithms on common uncertain RL tasks.

%% file: main.tex
%%%%%%%%%%%%%%%%%%%%%%%%%%%%%%%%%%%
% Introduction
\section{Introduction}

As Reinforcement Learning (RL) algorithms become increasingly competitive, attention is being shifted from toy and benchmark environments towards realistic everyday problems. 
However, real-world applications are never fully observable, resulting in various sources of uncertainty: imperfect actuators and sensors, miscalibration, user mental state, system wear and tear, etc. 
These uncertainties directly impact the policies as they cause stochasticity in the dynamics,  rewards, actions as well as observations \cite{cassandra1998survey, dulac2020empirical}.
In this work, we highlight that when considering commonly-used uncertain settings, the final policy of a standard RL algorithm may have large variations in its performance, i.e. low reproducibility.
We argue that evaluating and accounting for policy reproducibility is a critical and understudied issue in RL.

\begin{figure}[t!]
\centering
    \includegraphics[width = \linewidth]{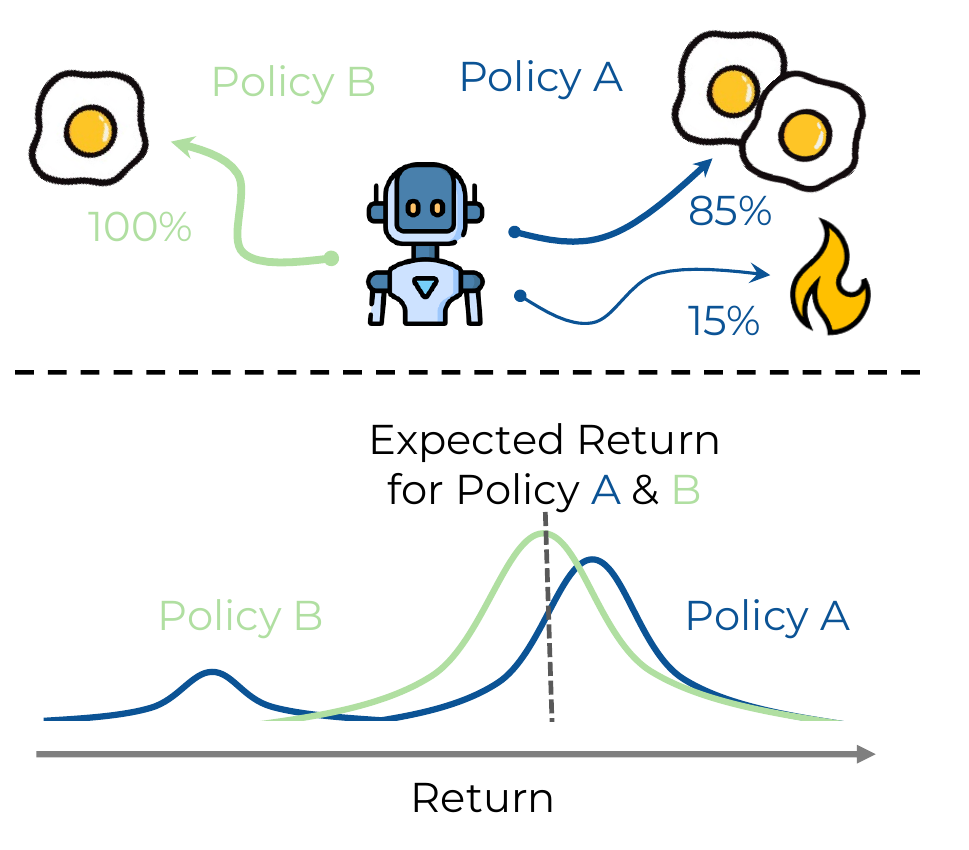}
    \caption{
        Illustration of the trade-off between policy reproducibility and performance. 
        Policy $A$ (blue) cooks the best-breakfast-ever-made 85\% of the time, but burns all the eggs the remaining 15\%, while Policy $B$ (green) consistently cooks a lower-quality breakfast. 
        On the distribution of returns (bottom), Policy $A$ and $B$ have the same expected return, highlighting the limitations of this commonly-used metric. 
    }
\label{fig:problem}
\end{figure}

As a motivating example, let's take the hypothetical example of a breakfast-cooking robot, trained using RL to cook the best expected breakfast every morning (see Figure~\ref{fig:problem}). 
A learnt robot Policy $A$ might get a really high expected return by cooking the best-breakfast-ever-made 85\% of the time, but burn all the eggs the remaining 15\% (blue Policy in Figure~\ref{fig:problem}).
This would result in finding burned eggs in the user's plates approximately one day a week. 
In this example, it is likely that the user would prefer a slightly lower-quality breakfast but consistently every day of the week, such as proposed by Policy $B$ (green Policy in Figure~\ref{fig:problem}). In other words, the user would likely prefer a more reproducible policy.
However, Policy $A$ and Policy $B$ both have the same expected return. 
More critically, the expected return metric does not leave any room to take into account this trade-off between quality and reproducibility, in order to choose the policy to deploy. 

More practically, consider the evaluation of a policy during training or as a result of an RL algorithm.
Common practice involves rolling out a policy $N$ (i.e. $N\sim200$) times and reporting the average return of the policy as the performance~\cite{agarwal2021deep}.
However, in the presence of uncertainty, these $N$ evaluation roll-outs would each lead to different performances, among which could even be failure cases. 
Thus, these $N$ roll-outs give an approximation of the distribution of returns of the policy (Figure~\ref{fig:problem}), that the average return metric fails to capture fully. 
% The same policy, which on average might perform well, can also have failure cases (see Figure~\ref{fig:sac_histogram}). 
% Thus, we propose an alternative evaluation metric that accounts for distribution-wide properties.
% Additionally, while the reward can at times capture behaviour, this is not always the case and policy behaviour can differ in uncertain domains despite achieving similar reward.
% Additionally, policy behaviour is important for interpretability in applied settings and hence capturing the reproducibility of behaviour is also critical as it can differ despite obtaining similar rewards.
In this work, we formalise the spread of this distribution as \textit{policy reproducibility}, which refers to the ability of a policy to obtain similar performance and behaviours when rolled out many times in uncertain domains. 
Here, we consider the standard \textit{in-distribution} case where the noise and uncertainty during evaluation have the same distribution as during training, but our proposed metrics could also be applied in \textit{out-of-distribution} cases such as sim-to-real transfer.

We argue that for RL to mature and successfully transition toward real-world applications, there is a need to move beyond reporting and evaluating just the expected (i.e. average) return, and start taking into account its trade-off with policy reproducibility.
Accordingly, we propose to use the Lower Confidence Bound (LCB), taken from Bayesian Optimisation~\cite{frazier2018tutorial}, as an alternative to the expected return. 
This metric provides practitioners with a preference parameter, which allows them to set the desired trade-off between expected performance and reproducibility. 
Additionally, the usual expected return can be recovered when this preference is set to zero.
We believe this point is crucial for the effective deployment of RL-generated policies in real-world settings. 

Prior work in RL has been concerned with the reproducibility of algorithms across seeds, and showed the importance of reporting statistically-meaningful metrics to ensure fair comparison of RL algorithms given a limited number of runs~\cite{henderson2018deep, agarwal2021deep}. 
Our work aims to further complement the evaluation of RL algorithms by focusing on reproducibility at the level of the policy. 

Our contributions are as follows: 
(1) we formalise the concept of policy reproducibility and propose the Mean Absolute Deviation (MAD) metric to quantify it, 
(2) we highlight a simple and easy-to-use alternative to the expected return that takes into account the trade-off between performance and reproducibility: the Lower Confidence Bound (LCB) of the return, 
%both in terms of return and behaviour. 
(3) we demonstrate the benefit of the LCB metric via a thorough experimental analysis of common RL algorithms across multiple domains,
%Our experiments find that algorithms such as Evolution Strategies (ES), which use parameter space noise and exploration for optimisation have inherent characteristics making learnt policies highly reproducible over value-based RL algorithms.
(4) we extend our evaluation procedures to also consider behavioural reproducibility.
\section{Background} \label{sec:background}

\subsection{Reinforcement Learning} \label{subsec:rl}

\subsubsection{Problem Statement}
In this work, we focus on solving Markov Decision Process (MDP) problems~\cite{sutton2018reinforcement}, in which an agent aims to maximise a reward signal. 
At each timestep $t$ of an MDP problem, the agent is in a state $s_t \in \mathcal{S}$, takes an action $a_t \in \mathcal{A}$, leading to a new state $s_{t+1} \in \mathcal{S}$. It then receives a reward $r_t = R(s_t, a_t, s_{t+1}) \in \mathbb{R}$ for this transition. 
The MDP is given as $<\mathcal{S}, \mathcal{A}, R, P>$ where $P:\mathcal{S}, \mathcal{A} \mapsto \mathcal{S}$ characterises the transitions dynamics.
% Most approaches also consider a discount factor $\gamma \in \mathbb{R}$, that allows to discount distant reward, given the cumulative discounted reward: $\sum_{t=0}^T \gamma^t r_{t+1}$.
In this work, we consider policies represented as a deep neural network $\pi_\theta$, parameterised by $\theta$.
These policies can either be deterministic, i.e. output a fix action, or stochastic, i.e. output a distribution over actions.
Algorithms aim to find parameters $\theta$ of the the policy $\pi_\theta \in \Pi, s_t \mapsto a_t$ that maximises the cumulative reward over episode of length $T$: $\sum_{t=0}^T r_t$. 
In this work, we consider and study two different families of algorithms used in RL.

\subsubsection{Value-based Methods} use the gradient with respect to timestep reward as a learning signal to find the optimal policy. 
To do so, they usually approximate the state-action value function $Q^\pi: \mathcal{S}, \mathcal{A} \mapsto \mathbb{R} ; s_t, a_t \mapsto \mathbb{E}_{\pi}\left(\sum_{t=0}^T \gamma^t r_t | s_t, a_t \right)$, which gives the expected discounted return for being in state $s_t$, taking action $a_t$ and following policy $\pi$.
$Q$ is usually approximated using a deep neural network with parameter $\phi$, referred to as the critic network $Q_\phi$. 
%The policy is also represented as a deep neural network, parameterised by $\theta$ and referred to as the actor $\pi_\theta$. 
%$Q_\phi$ is optimised to approximate $Q$, while $\pi_\theta$ learns to maximise $Q_\phi$ following the policy gradient theorem \cite{silver2014deterministic, sutton2018reinforcement}.
The policy $\pi_\theta$ learns to maximise $Q_\phi$ following the policy gradient theorem~\cite{sutton2018reinforcement}.
We consider two popular algorithms: TD3~\cite{fujimoto2018addressing}, which uses deterministic policies, and SAC~\cite{haarnoja2018soft}, which uses stochastic policies.

\subsubsection{Evolution Strategies}~\label{sec:es}
(ES) use as a learning signal the (empirical) gradient with respect to the cumulative reward over the episode~\cite{hansen2006cma, wierstra2014natural}. 
ES maintain a parameterised search distribution over policy parameters $\theta$. 
%Evolution Strategies~\cite{hansen2006cma, wierstra2014natural} are black-box optimisation algorithms which consist of a parameterized search distribution over solutions $\theta$. 
% Parameters of the search distribution are then updated by moving in the direction of higher fitness solutions.
They update this distribution in the direction of higher cumulative reward by following the natural gradient obtained through sampling in the solution space from the search distribution.
ES algorithms come in many variants and are classified based on the distribution parameterization and the update procedure.
In this paper, we focus on the OpenAI-ES~\cite{salimans2017evolution} variant, shown to work well for our problem setting, which we refer to as just ES for simplicity.
% ES uses an isotropic Gaussian $\mathcal{N}(\theta_\mu, \sigma^2I)$ to represent the search distribution, where $\mu$ is the mean and $\sigma^2$ is the variance.
% A population of $N$ solutions are sampled $\theta_i = \theta_\mu + \sigma\epsilon_i$ where $\epsilon_i \sim \mathcal{N}(0, I)$. from this distribution at every iteration of the algorithm. Each sample $\theta_i$ is then evaluated on the objective function $F(\theta_i)$ to obtain a fitness score. The mean $\mu$ of the distribution can then be updated with the search gradient $\theta_{\mu}^{t+1} \gets \theta_{\mu}^t + \alpha\nabla E_{\theta\sim\mathcal{N}(\theta_\mu, \sigma)}[F(\theta)]$ which is estimated according to:

% \begin{align}
%     \nabla E_{\theta\sim\mathcal{N}(\theta_\mu, \sigma)}[F(\theta)] \approx \frac{1}{N\sigma}\sum{F(\theta_i)\epsilon_i}
% \end{align}

%%%%%%%%%%%%%%%%%%%%%%%%%%%%%%%%%%%
% Quick survey of previous work
\subsection{Uncertainty} \label{subsec:uncertainty}

\begin{figure}[h]
\centering
    \includegraphics[width = \linewidth]{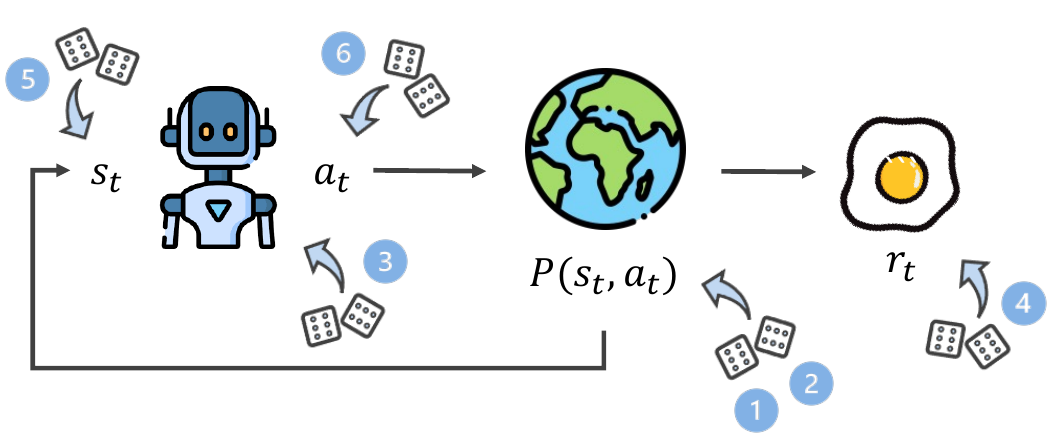}
    \caption{
        Illustration of the different uncertainties that can be applied within the RL evaluation setting: (1) stochastic dynamics, (2) random initialisation, (3) parameter-space noise, (4) reward noise, (5) observation noise and (6) action noise.
    }
\label{fig:uncertainty}
\end{figure}

We consider environments with different types of noise and stochasticity, referred to more generally as uncertain. 
As the issue of reproducibility only arises in such uncertain settings, this section provides an overview of the types and sources of uncertainty illustrated in Figure~\ref{fig:uncertainty}. 
For each type, we give the objective maximised by the RL agent, for example, when there is no uncertainty in the environment: 

\begin{center}
    \small
    $J(\theta) = \mbox{E}_{s_0, P, \pi_\theta}\left(\sum_{t=0}^T  R(s_t, a_t, s_{t+1}) \right)$
\end{center} 

\subsubsection{Stochastic Dynamics}

Stochastic dynamics has been long studied in the field of RL. The transition matrix $P$ defined in Section~\ref{subsec:rl} can be stochastic, where the same action from the same state can lead to different next states (Fig.~\ref{fig:uncertainty}.1): 

\begin{center} 
    \small
    $J(\theta) = \mbox{E}_{s_0, \tilde{T}, \pi_\theta}\left[\sum_{t=0}^T R(s_t, a_t, s_{t+1}) \right]$

    with $\tilde{T}$ such that $\exists (i, j, k), \tilde{T}_{ij} > 0$ and $\tilde{T}_{ik} > 0$
\end{center} 

Prior works in this setting focus on the issue of bias and early commitment \cite{fox2015taming, hasselt2010double, azar2011speedy}, where policies encountering lucky states early in learning spent lots of effort later unlearning biased estimates.

\subsubsection{Random Initialisation}
Most environments in RL literature have random initialisation \cite{tassa2018deepmind}  (Fig.~\ref{fig:uncertainty}.2): 

\begin{center}
    \small
    $J(\theta) = \mbox{E}_{s_0 + \epsilon, T, \pi_\theta}\left[\sum_{t=0}^T R(s_t, a_t, s_{t+1}) \right] \texttt{, } \epsilon \sim \mathcal{N}(0, \sigma)$
\end{center}

Unlike the previous one, this noise does not impact the learning capabilities of RL algorithms. On the contrary, it often proves useful by enhancing exploration.

\subsubsection{Parameter-space Noise}
Literature in ES and, more generally, in Evolutionary Algorithms (EA) also commonly consider noise applied directly on the policy parameters $\theta$ \cite{jin2005evolutionary, lehman2018more} (Fig.~\ref{fig:uncertainty}.3):

\begin{center}
    \small
    $J(\theta) = \mbox{E}_{s_0, T, \pi_{\theta + \epsilon}}\left[\sum_{t=0}^T R(s_t, a_t, s_{t+1}) \right] \texttt{, } \epsilon \sim \mathcal{N}(0, \sigma)$
\end{center}

ES algorithms which optimise based on perturbations in the parameter space have demonstrated greater performance in this setting than common RL methods~\cite{lehman2018more}.
% Interestingly, when considering feedforward neural network policies, this is connected to uncertainty on the policy input.

\subsubsection{Reward Noise}

Another type of uncertainty considers additive noise directly on the reward value~\cite{everitt2017reinforcement, wang2020reinforcement, romoff2018reward}, coming for example from imperfect sensors (Fig.~\ref{fig:uncertainty}.4). 
Such noise directly impacts the optimisation and might lead to learning sub-optimal policies.
Interestingly, this issue is also widely studied in EA and ES~\cite{jin2005evolutionary, rakshit2017noisy}. 

\begin{center}
    \small
    $J(\theta) = \mbox{E}_{s_0, T, \pi_{\theta}}\left[\sum_{t=0}^T R(s_t, a_t, s_{t+1})  + \epsilon_t \right] \texttt{, } \epsilon_t \sim \mathcal{N}(0, \sigma_t)$
\end{center}

Reward noise is a particular case of the reward corruption problem defined in \citet{everitt2017reinforcement}. %, or perturbed reward problem defined in \citet{wang2020reinforcement}. 
However, while the noisy reward problem can be alleviated, \citet{everitt2017reinforcement} derived a no-free lunch theorem for reward corruption.

\subsubsection{Observation Noise}

One can also consider noise on the state of the environment \cite{dulac2020empirical}, coming for example from imperfect sensors (Fig.~\ref{fig:uncertainty}.5):  

\begin{center}
    \small
    $J(\theta) = \mbox{E}_{s_0 + \epsilon_0, T, \tilde{\pi_{\theta}}}\left[\sum_{t=0}^T R(s_t + \epsilon_t, a_t, s_{t+1}  + \epsilon_{t+1}) \right]$ \\ $\epsilon_0 \sim \mathcal{N}(0, \sigma_0) \texttt{, } \epsilon_t \sim \mathcal{N}(0, \sigma_t) \texttt{, }  \epsilon_{t+1} \sim \mathcal{N}(0, \sigma_{t+1})$
\end{center}

Such noise is often considered in work interested in generalisation where it is integrated by the user as a way to enforce robust policies, for example in domain randomisation \cite{akkaya2019solving}. 
While this case is slightly different from the one considered here, it still provides interesting insights into the convergence of policies in such domains. 

\subsubsection{Action Noise}
One can also consider action noise, coming for example, from imperfect actuators (Fig.~\ref{fig:uncertainty}.6) and acts similar to observation noise~\cite{dulac2020empirical}:

\begin{center}
    \small
    $J(\theta) = \mbox{E}_{s_0, T, \tilde{\pi_{\theta}}}\left[\sum_{t=0}^T R(s_t, a_t + \epsilon, s_{t+1}) \right] \texttt{, } \epsilon \sim \mathcal{N}(0, \sigma)$
\end{center}

% Similar to observation noise, action noise is also often added to improve policies generalisation, providing interesting insights on convergence guarantees in this setting.

%%%%%%%%%%%%%%%%%%%%%%%%%%%%%%%%%%%
% Reproduciblity
\section{Accounting for Policy Reproducibility}

\begin{figure}[h]
\centering
    \includegraphics[width = \linewidth]{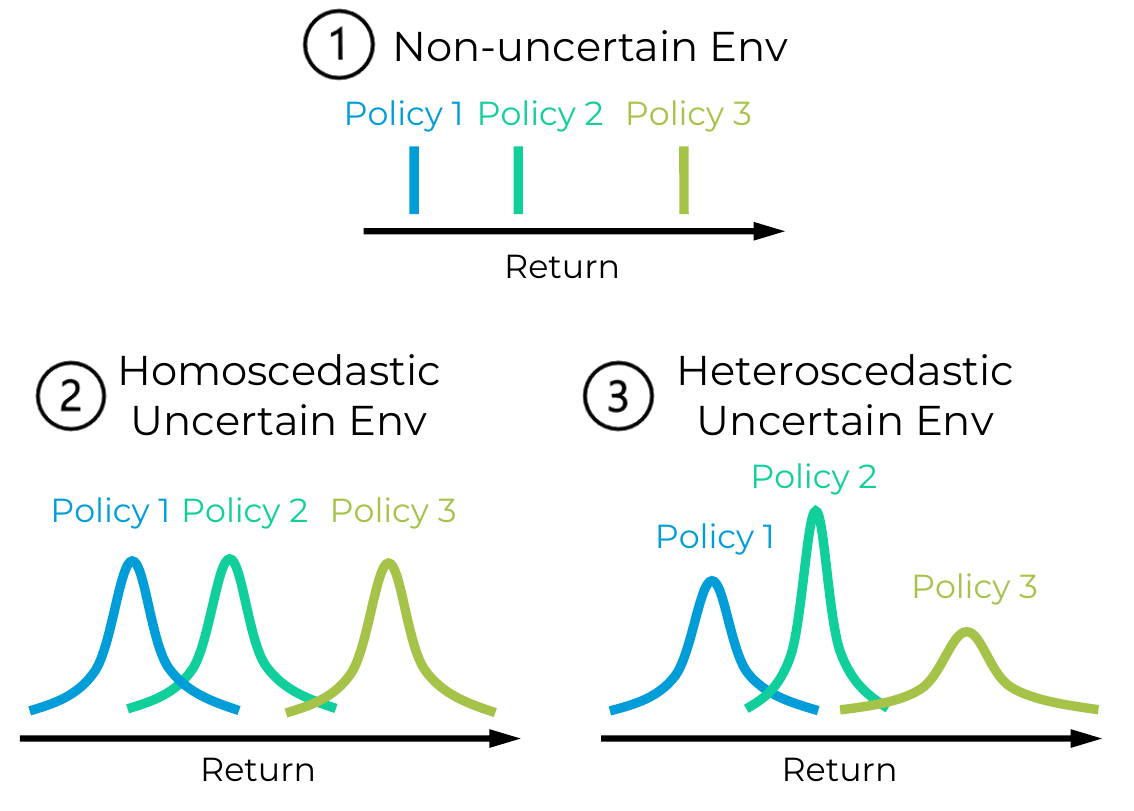}
    \caption{
        In non-uncertain environments (1), each policy has a fixed return; while in uncertain environments (2, 3), policies have distributions over possible returns. In homoscedastic uncertain environments (2) this distribution is the same for every policy, while in heteroscedastic uncertain environments (3), each policy can take different distribution parameters. The trade-off between policy reproducibility and performance only arises in heteroscedastic uncertain environments (3). 
    }
\label{fig:uncertain_env}
\end{figure}

\subsection{Uncertain Environments}
%We first refine the definition of uncertain environments. 
Section~\ref{subsec:uncertainty} provided an overview of the different types of uncertainty that can apply in an environment, but they might not all have the same impact.
\textit{Non-uncertain environments} refer to environments where each policy is attributed a single return value in a deterministic manner: every roll-out of a policy leads to the exact same return value (case (1) in Figure~\ref{fig:uncertain_env}). 
This occurs in the absence of uncertainties, or in the presence of uncertainties that do not propagate to the return, for example, uncertainties that impact only unreachable states. 
% These cases would lead to the same as non-uncertain environments.
On the contrary, \textit{uncertain environments} refer to environments where policies have distributions over return values: the same policy evaluated multiple times gets different return values from one roll-out to another (case (2) and (3) in Figure~\ref{fig:uncertain_env}). 
% This is what we refer to as uncertain environment.

\subsubsection{Homo/Hetero-scedastic Uncertain Environments.}
Uncertain environments can be further divided into two categories.
\textit{Homoscedastic uncertain environments} refers to environments where the uncertainty impacts the returns of all policies in the same way, while in \textit{heteroscedastic uncertain environments}, it depends on the policy considered.
In other words, both types of environments are uncertain, but in homoscedastic ones, the return of all policies have the same distribution shape (case (2) in Figure~\ref{fig:uncertain_env}) while in heteroscedastic ones, the return distributions of policies can be different (case (3) in Figure~\ref{fig:uncertain_env}). 
An example of a homoscedastic uncertain environment would be an environment where the only source of noise is a noisy sensor that adds a constant white noise to the reward.
In this case, all policies would get the same return distribution spread, no matter their performance. 
An example of a heteroscedastic uncertain environment would be initial state noise as this uncertainty would propagate differently through the episode and can lead to a different return distribution for each policy.
We would like to emphasise that homoscedasticity and heteroscedasticity qualify environments and not policies as it is a property of the search space as a whole.

\subsection{Policy Reproducibility} \label{subsec:quant_rep}

Consider a heteroscedastic uncertain environment and a policy $\pi$. 
The performance $P_{\pi}$ of $\pi$ would usually be quantified using its expected return, but this single number does not carry any information on the spread of the distribution.
We define the \textbf{policy reproducibility} $\sigma_{\pi}$ as the statistical dispersion of the return distribution of $\pi$.
% We propose to consider the spread, or statistical dispersion, of the distribution as a second dimension of interest, and we refer to it as the reproducibility $\sigma_{\pi}$.
Both $P_{\pi}$ and $\sigma_{\pi}$ can be quantified using multiple metrics, for example, the mean or mode for $P_{\pi}$, and the standard deviation or entropy for $\sigma_{\pi}$. 
% In the next section, we propose one quantification for $\sigma_{\pi}$ that is robust to outliers.
% We propose two possible quantifications for $\sigma_{\pi}$ in the next section. 

\begin{figure}[h]
\centering
    \includegraphics[width = 0.95\linewidth]{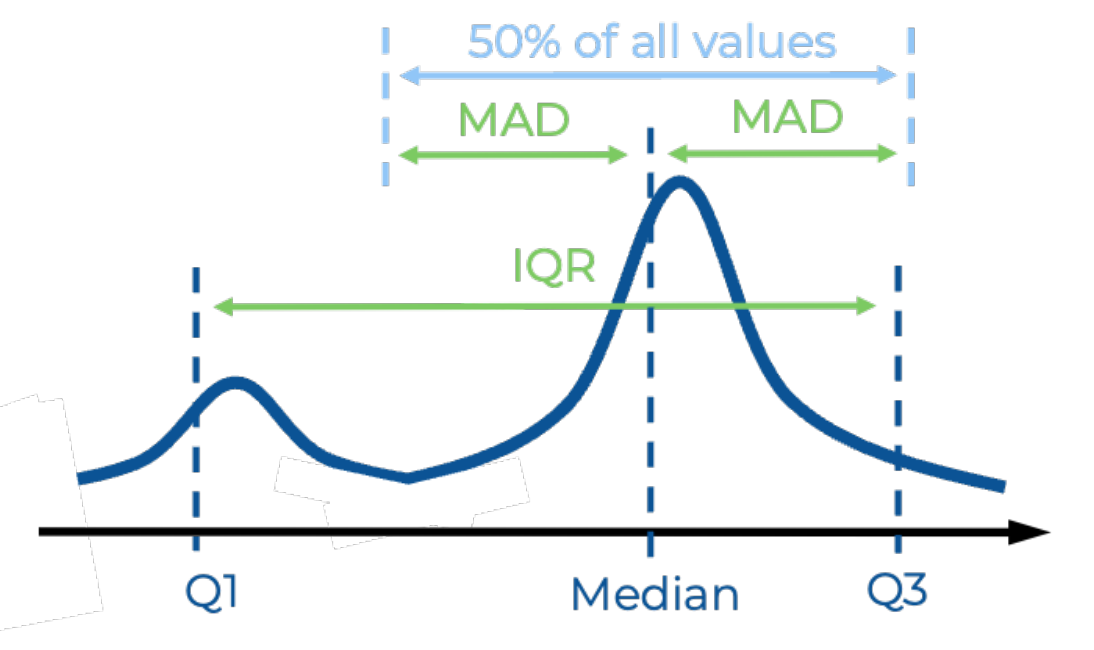}
    \caption{
    Illustration of the MAD and IQR metrics that quantify policy reproducibility for a given return distribution. 
    IQR corresponds to the distance between first and third quartiles, while MAD corresponds to the median distance to the median of the distribution. 
    As a consequence, half the sampled evaluations are closer to the median than the MAD, and half are further away, as illustrated in the figure.
    %Both quantify the reproducibility, while the average return quantify performance.
    %Common distribution of returns when evaluating a policy $N\sim200$ times in uncertain domains.
    %The average return does not capture aspects of policy performance which are important for moving RL as a field from research to applied settings where reproducibility is important. The worst return and return variance can provide additional information on the reproducibility.
    }
\label{fig:new_metrics}
\end{figure}

\subsubsection{Robust Quantification of Policy Reproducibility.}
$\sigma_{\pi}$ can be quantified using any common measure of statistical dispersion, such as the standard deviation. 
However, common evaluation procedures rely on $N$ evaluations of a policy $\pi$ to approximate both $P_{\pi}$ and $\sigma_{\pi}$.
As $N$ is usually chosen quite low, this procedure is likely to contain outlier evaluations that can greatly impact the approximation of $\sigma_{\pi}$.
Thus, we propose using the Median Absolute Deviation (MAD) estimator, known for its robustness to outliers~\cite{wilcox2011introduction} to quantify policy reproducibility $\sigma_\pi$. 
We denote $R^N = (R_i)_{i \in [0, N]}$ as the set of the $N$ evaluations of the return of a policy, also illustrated in Figure~\ref{fig:new_metrics}:
\begin{center}
    $MAD_R = median(|R^N - median(R^N)|)$
\end{center}

While we choose to focus on the MAD metric, an alternative dispersion measure also known for its robustness to outliers would be the Inter-Quartile Range (IQR)~\cite{wilcox2011introduction} which gives a wider representation of the return distribution and a better indication of worst-case returns.
The return IQR (see Figure~\ref{fig:new_metrics}) is computed as $IQR_R = Q_3(R^N) - Q_1(R^N) \text{, where } Q_i \text{ is the } i^{th} \text{ quartile}$.

\subsubsection{Lower Confidence Bound for Policy Comparison.}
The MAD defined in the previous section, allows the reproducibility of a policy $\sigma_{\pi}$ to be quantified. 
However, the crucial question remains how to reliably compare RL policies while taking into account the trade-off between performance $P_{\pi}$ and reproducibility $\sigma_{\pi}$. 
% The expected return has two major limitations: first, it does not carry any information on the reproducibility so two policies with same expected return could have really different distribution properties (see Figure~\ref{fig:problem}). Secondly, it does not allow the practitioners to set their preferences in terms of performance versus reproducibility.
% To overcome these issues, 
% To account for policy reproducibility when comparing policies resulting from RL algorithms, a new metric that respects the trade-off between performance $P_{\pi}$ and reproducibility $\sigma_{\pi}$ is required. 
To this end, we propose a simple alternative to the commonly used expected return: the lower-confidence bound (LCB)~\cite{frazier2018tutorial}, given as:
$$LCB(\pi) = P_{\pi} - \alpha \text{ } \sigma_{\pi}$$
where $\alpha \in \mbox{R}^+$ is a user-chosen parameter that allows weighting the relative importance of the performance and reproducibility. 
When the performance and the reproducibility values are normalised within the same interval, $\alpha$ can be interpreted as trading $\alpha \%$ of performance for $\alpha \%$ of reproducibility. 
As the LCB score is only computed after the runs
(during policy evaluation), all the statistics needed for choosing the value of $\alpha$ are available. 
Additionally, $\alpha$ does not constitute a learning hyper-parameter, as it is only used when evaluating and the learning algorithms are agnostic to it. 
%$\alpha$ is only used when evaluating and comparing policies resulting from algorithms.

In the following, we choose $P_{\pi}$ as the expected return over $N$ evaluations and $\sigma_{\pi}$ as the MAD over these $N$ evaluations. 
This decision allows the conventional expected return to be recovered when $\alpha = 0$, i.e. performance only.
For robustness to outlier evaluations when $N$ is low, we recommend using robust estimators such as median for $P_{\pi}$.
We also report comparisons using the median for $P_{\pi}$ and standard deviation and IQR for $\sigma_{\pi}$ in Appendix~\ref{app:lcb}.

% For all these quantifications, the user can define a preference parameter $\alpha$ that weights the relative importance of $R_{\pi}$ and $\sigma_{\pi}$ in selecting the final policy to deploy in the environment. 

%%%%%%%%%%%%%%%%%%%%%%%%%%%%%%%%%%%
% Results
\begin{figure}[t!]
\centering
    \includegraphics[width = \linewidth]{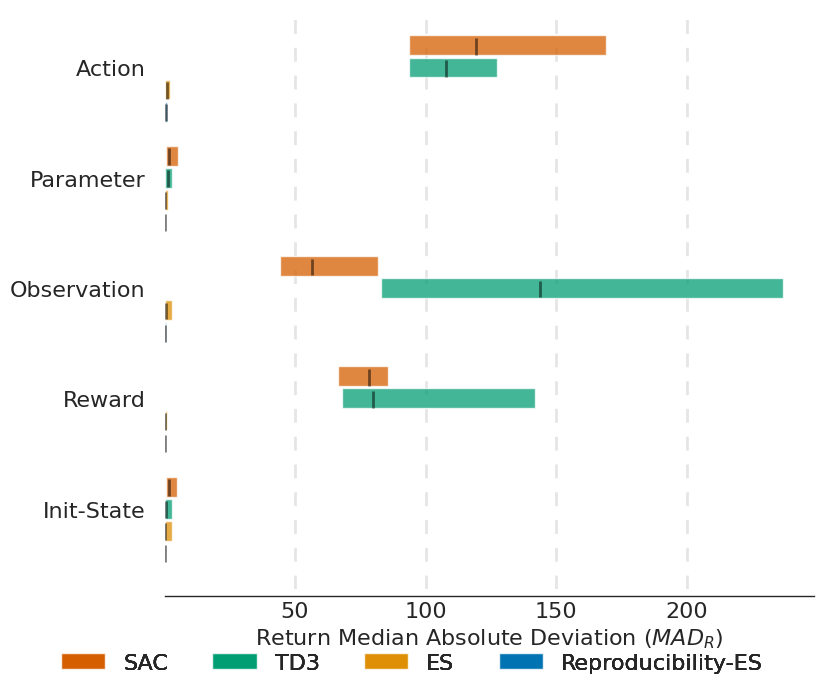}
    \caption{
        MAD scores of final policies in the Ant environment. y-axis is the type of uncertainty present in the environment.  
        We report the IQM and the CIs across 10 seeds.
        The lower the MAD score the more reproducible the policy.
    }
\label{fig:results_MR_WR}
\end{figure}

% \section{Evaluating Policy Reproducibility of existing algorithms}
\section{Experiments}

In this section, we evaluate policies resulting from commonly used RL algorithms using the introduced MAD metric, and corresponding LCB measures.
% \subsection{Baselines}
We benchmark the reproducibility of resulting policies from SAC, TD3 and ES as detailed in Section~\ref{sec:background}. For SAC, action sampling is removed during evaluation for fair comparison.
We also introduce an ES variant that explicitly optimises for reproducibility which we refer to as Reproducibility ES (R-ES).
R-ES uses ES to optimise for a weighted sum of performance and reproducibility; where performance is computed as the average return over $32$ re-evaluations, and the reproducibility as the standard deviation of these re-evaluations.
% In R-ES, each sample used to compute the approximate gradient is evaluated $32$ times, and the variance of return across these re-evaluations is used as an objective to be minimised.
% This term is treated as a weighted sum with the return of the sample.
% A full list of hyper-parameters for each algorithm are available in 
Algorithm hyper-parameters are provided in Appendix~\ref{app:hyperparams}.

We conduct experiments on common continuous control tasks (Ant and HalfCheetah) using the Brax~\cite{freeman2021brax} simulator.
A policy is evaluated by rolling it out $N=256$ times in the environment, providing a distribution of returns and trajectories.
The reproducibility quantification MAD and the LCB metric presented in Section~\ref{subsec:quant_rep} are then computed based on these roll-outs.
We report results in a consistent manner across metrics and environments, in the form of the inter-quartile mean (IQM) and stratified bootstrap confidence interval (cis) across seeds using the rliable library~\cite{agarwal2021deep}.
We report replications of results for each algorithm and environment with $10$ seeds.

\input{figures/mr_wr_table}

\subsection{Effect of Types of Noise}
We first conduct experiments to study the effect of a variety of noise and uncertainty on reproducibility.
It is important to note that the same distribution of noise is applied during training and during evaluation.
Figure~\ref{fig:results_MR_WR} and Table~\ref{tab:lcb_results} show the MAD scores, quantifying the reproducibility of the resulting policies.
As a base experiment, we consider the standard RL setting with random initial states and stochastic transitions.
We then independently consider the noise on the action, observation, reward and parameters, on top of this base setting.
From the algorithms considered, policies obtained from SAC are observed to obtain a higher MAD, demonstrating a large spread in return values (low reproducibility). 
% This is also reflected in the WR which shows large values, indicating the presence of low-return (failed) rollouts during the evaluation.
Interestingly, ES-based approaches result in more reproducible policies in all the cases of noise, evident from the lower MAD values. Policies learnt using ES are consistently more reproducible, and explicitly maximising for reproducible policy with the simple R-ES method also shows some additional improvements.
% Our experiments show this is also true when using the IQR metric.
This trend is also true for the IQR metric reported in Appendix~\ref{app:wr}.

\begin{figure}[t!]
\centering
    \includegraphics[width = \linewidth]{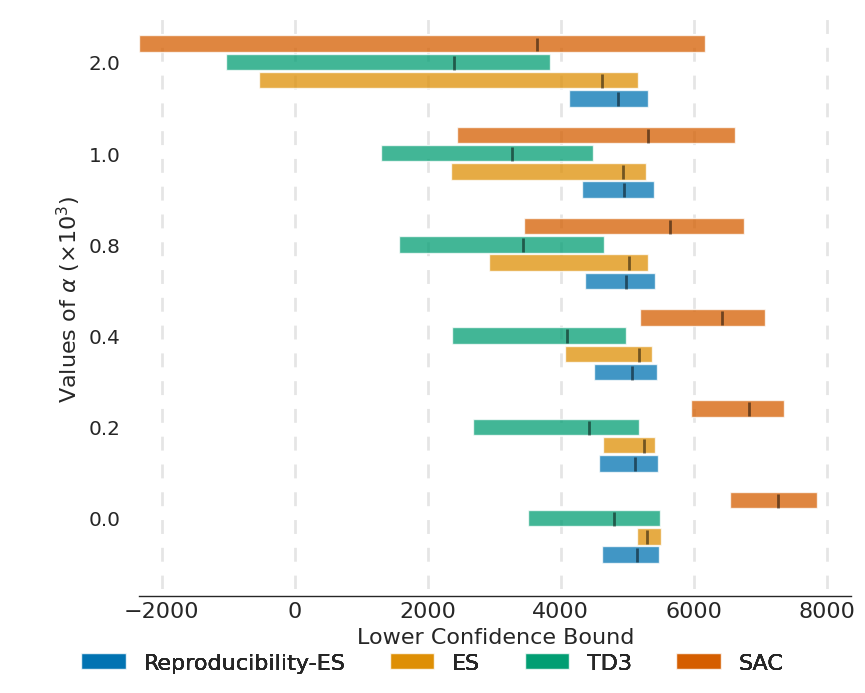}
    \caption{
        LCB comparison on the Ant with Init-State noise. 
        y-axis corresponds to varying values of $\alpha$, the trade-off between performance and reproducibility. We report the IQM and the CIs across 10 seeds. The higher the LCB score the better.
    }
\label{fig:results_LCB}
\end{figure}

\input{figures/lcb_table}

\subsection{Evaluating Performance and Reproducibility}
The increase in reproducibility observed for ES methods seems to come at a performance cost, observed from the lower expected return when $\alpha=0$ (Fig.~\ref{fig:results_LCB}).
To better understand the performance-reproducibility trade-off and make policy comparisons that account for reproducibility, Figure~\ref{fig:results_LCB} and Table~\ref{tab:lcb_results} show the use of the proposed LCB metric. 
Increasing values of $\alpha$ indicates increasing consideration for policy reproducibility.
As previously explained, the conventionally used expected return would be recovered when $\alpha=0$.
As $\alpha$ is increased, there is a clear and evident tradeoff between policy reproducibility and performance.
We observe that the LCB of a policy that is reproducible (as a result of ES algorithms) is relatively unaffected by the increase in $\alpha$, while there can be a significant drop in LCB score when a policy is not reproducible due to a large distribution spread (SAC and TD3).
In the case of the baseline algorithms considered, we hypothesise that this is due to optimisation properties in ES that come with perturbations in the parameter space, and optimising for the average reward of the population obtained from these perturbations~\cite{lehman2018more}.
This implicitly leads to more reproducible policies but comes at the cost of not being able to move to less reproducible parts of the search space where higher performance is possible.

Overall, our results demonstrate the benefit of LCB to account for both policy reproducibility and performance.
More practically, the LCB allows practitioners to account for this reproducibility using $\alpha$, given the tolerance in corresponding applications.
Interestingly, the performance/reproducibility trade-off can also be studied through the lens of multi-objective optimisation, and a Pareto plot can be used to compare algorithms (see Appendix~\ref{app:pareto}).

\section{Extension to Behaviour Reproducibility} \label{sec:behaviour}

\begin{figure}[h!]
\centering
    \includegraphics[width = 0.97\linewidth]{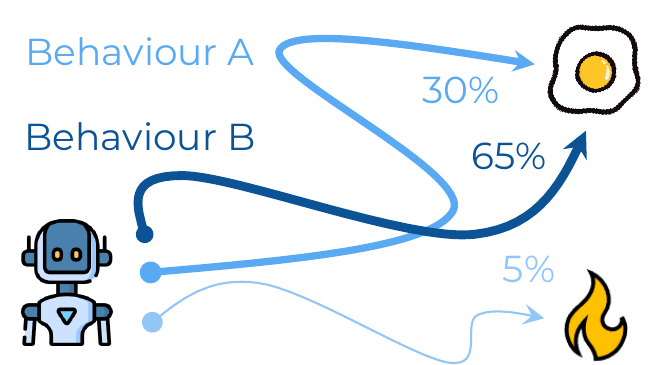}
    \caption{Illustration of the \textit{behaviour reproducibility} in uncertain domains.
    \textit{Behaviour reproducibility} refers to the different behaviours exhibited by a policy, agnostic to the return obtained.
    Behaviour $A$ and $B$ can have similar performance.
    }
\label{fig:behaviour_problem}
\end{figure}

While the reward can at times capture behaviour, this is not always the case and policy behaviour can differ in uncertain domains despite achieving similar returns (see Figure~\ref{fig:behaviour_problem}).
Using the example of the breakfast-cooking robot from the introduction again, consider the robot gathering ingredients. 
If the robot always uses the same trajectory to do so, the user might get used to and expect a certain behaviour and adapt its own habit in the kitchen accordingly. 
If the robot randomly changes its trajectory in the exact same setting despite accomplishing the task, this might become a risk for the user. 
Hence, behavioural reproducibility is essential for policy interpretability and can become critical in applied settings.
In this section, we extend our formalisation and metrics for performance reproducibility to behaviour reproducibility. 

\subsubsection{Behavioural Representation.} To evaluate the behaviour reproducibility, we use two different measures of behavioural representation $\mathbf{b}$.
The first measure is derived from novelty search~\cite{lehman2011abandoning} and Quality-Diversity~\cite{pugh2016quality, chatzilygeroudis2021quality} literature, where a \textit{behavioural descriptor} is defined for each solution or policy to characterise a policy with respect to another.
For example, the average time each foot is in contact with the ground to characterise the gait of a legged robot~\cite{cully2015robots}, or the final position of an agent in a maze~\cite{lehman2011abandoning}.
Behaviour descriptors have already been used to quantify behaviour reproducibility for Quality-Diversity~\cite{flageat2023uncertain}.
Next, we also consider the \textit{state marginal} distribution of the different roll-outs of the policy as a representation of the behaviour.
While we use these two measures, other forms of behavioural representation can also be used such as action embeddings~\cite{parker2020effective} or learnt state-trajectory encodings~\cite{cully2019autonomous, lynch2020learning}.

\subsubsection{Quantifying Behaviour Reproducibility}
Unlike the return, behaviour representations are multidimensional and their absolute values carry no effective meaning (such as "larger is better"). 
Thus, the MAD metric defined in Section~\ref{subsec:quant_rep} cannot be applied out-of-the-box. 
We propose to apply them as second-order metrics on the distance between behaviour representations.
In other words, denoting $B^N = (b_i)_{i \in [0, N]}$ as the set of the $N$ evaluations of the behaviour of a policy, $D^N = (distance(B^N, B^N))$ is the set of all the two-by-two distances of $B^N$, for any distance function. 
We define the behavioural MAD as:
$MAD_{B} = median(|D - median(D)|)$.

\begin{figure}[t!] 
\centering
    \includegraphics[width = \linewidth]{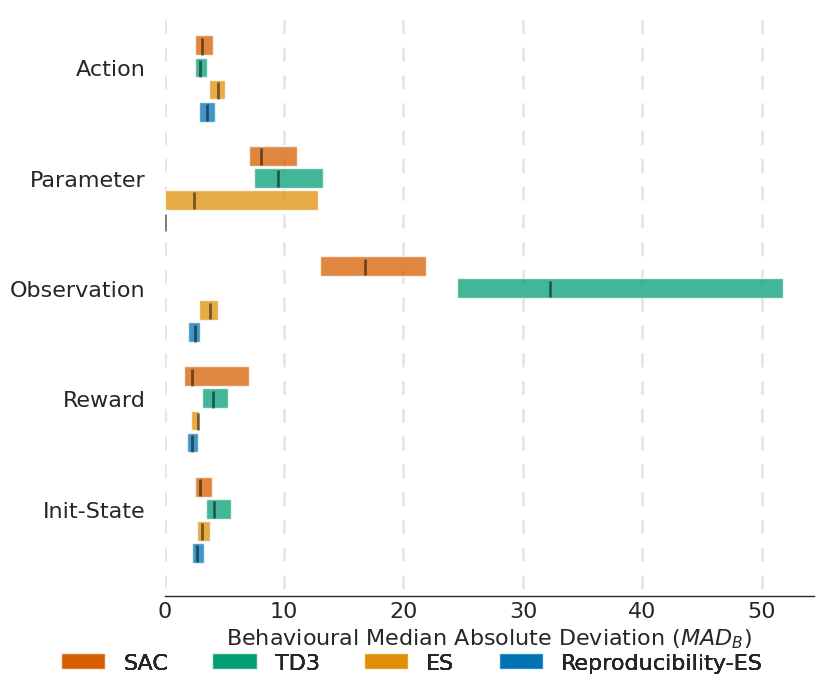}
    \caption{
    Behavioural MAD scores across types of uncertainty present in the Ant environment.  
    We report the IQM and the CIs across 10 seeds. 
    The lower the MAD score the more reproducible the behaviour.
    }
\label{fig:results_MBR_WBR}
\end{figure}

\subsubsection{Experiments}
Figure~\ref{fig:results_MBR_WBR} shows the behavioural MAD when considering the behavioural descriptor.
The descriptor in this case corresponds to the average foot contact of the robot across the episode~\cite{cully2015robots}.
We observe similar trends when comparing the algorithms in the case of behavioural reproducibility, where R-ES consistently performs well despite optimising only for return reproducibility.
However, the behavioural reproducibility of ES is less strong across other tasks (see Appendix~\ref{app:bd}).
We hypothesise that the computation of behavioural representations are more sensitive than return values, causing uncertainty and errors to have a larger effect on behaviours.
Similar results are observed when using state-marginal representation (see Appendix~\ref{app:bd}).
%The results for the state-marginal representation are reported in Appendix~\ref{app:bd}.

\section{Related Work}

\subsection{Safe and Robust RL} \label{subsec:robust_safe_rl}

\subsubsection{Safe RL.} 
Many real-world applications such as robotics or autonomous driving require guarantees that policies will not lead to unsafe situations such as damages or collisions. 
Safe RL~\cite{garcia2015comprehensive, brunke2022safe, gu2022review} focuses on such safety-critical applications and aims to solve them using RL. 
To do so, it expresses the unsafe regions using a set of constraints that need to be satisfied by the RL agent to ensure safe operation. 
In many applications, these constrains do not necessarily have to be hard constraints, and only need to be satisfied with a high probability.
Our work improves evaluation procedures in the general RL setting and not necessarily in constrained setups. 
While behaviour reproducibility shares some similarities with stability, which refers to the boundedness of the system output or state in Safe RL \cite{brunke2022safe}, reproducibility only quantifies statistical dispersion and does not enforce constraints. 
Our proposed metrics are complementary and can also be applied in the Safe RL setup as policies can have low reproducibility while satisfying required safety constraints. 

\subsubsection{Robust RL.}
Robust RL~\cite{morimoto2005robust, moos2022robust, chen2020overview} focuses on learning policies that, while trained on a given uncertainty distribution, are able to maintain performance when applied to a distinct test uncertainty distribution.
This property makes them better at handling sim-to-real gaps or facing new and unplanned situations. 
A common approach in robust RL is to re-formulate the RL problem as a two-player game, where an adversary aims to deteriorate the policy performance by manipulating the environment~\cite{morimoto2005robust, moos2022robust, chen2020overview}. 
\citet{moos2022robust} proposed a classification of robust RL approach based on the part of environment targeted by the adversary, that draws an interesting parallel with the classification of uncertainties from Section~\ref{subsec:uncertainty}.
%This classification includes: transition dynamics uncertainty, including reward~\cite{nilim2005robust}, action uncertainty~\cite{tessler2019action}, observation uncertainty~\cite{pinto2017robust}, and finally disturbance addition \cite{morimoto2005robust}. 
Similar to Robust RL, our work is concerned with the evaluation and the maintenance of the performance of policies under uncertainty.
Closest to our work, ~\citet{xu2006robustness} explores the trade-off between performance and robustness.
However, our work defines reproducibility more generally for any policy without the need for robustness algorithms and the robust RL setting.
We are interested in improving existing evaluation procedures and metrics more generally across RL.
% Additionally, Robust RL obtains reproducibility as a by-product of we highlight the trade-off between reproducibility and performance.
% By acknowledging that maximising reproducibility explicitly (as done in Robust RL) is problem dependant,  we provide all RL practitioners a general metric to account for reproducibility given the tolerance for this.
% Additionally, our work applies to the standard RL setting where the test distribution is the same as the training distribution. 

%In this work, we highlight the importance of accounting for policy reproducibility, but we do not advocate for maximising it explicitly in every case, nor for maximising it over performance. 
%We provide the user with tools that allow them to set their own trade-off between performance and reproducibility and we acknowledge that different applications might require different trade-off. 
%From this point of view, our work shares motivation with the work done by~\citet{xu2006robustness} on trade-off between performance and robustness.
%Additionally, our work applies broadly to the general RL setting, without any assumption on training and testing uncertainty distributions. 
%Many robust RL approaches implicitly tackle the issue of policy reproducibility, which can even be viewed as a sub-problem in some cases. 
%However, our results highlight that this issue is largely understudied in the simple case of similar training and testing distributions. 

\subsection{Accounting for Value Distribution in RL}

\subsubsection{Distributional RL.} 
While many RL works learn the expectation of the value function, Distributional RL~\cite{bellemare2023distributional, bellemare2017distributional} proposes to learn the full value distribution. 
Learning such a distribution improves the performance of standard RL approaches as it provides a more stable learning target.
In comparison, our work considers return distributions of a policy, meaning episode-based and not timestep-based distributions. 
Additionally, these distributions are used to improve evaluation procedures and comparisons between policies and do not affect the learning algorithms in any way. 

\subsubsection{Using return variance for risk-sensitive RL.}  
Risk-sensitive RL approaches are aware that optimal policies may perform poorly in some cases, causing a risk of large negative outcomes. 
Thus, this body of work accounts for the variance of the expected return from the current state (value function), in order to minimise this risk~\cite{heger1994consideration, coraluppi1999risk}. 
Similar to Distributional RL, Risk-sensitive RL uses information on the distributions of time-step quantities (value functions) to improve learning while our work focuses on evaluation procedures and episodic return distributions. 

% \subsubsection{Imitation Learning.}
% The field of imitation learning~\cite{schaal1996learning, argall2009survey} involves learning a policy from expert demonstrations.
% In particular, Behavioural Cloning (BC)~\cite{bain1995framework, giusti2015machine, bojarski2016end} learns from a dataset of expert trajectories or transitions. 
% BC approaches suffer from errors in the policy and uncertainties in the environment as it leads the policy to diverge. 
% Our work considers the RL setting which does not necessarily assume access to expert demonstrations. 
% Naturally, maximising reproducibility using imitation learning techniques after a good policy is learnt is a possible alternative.
% Additionally, our proposed metrics can also be similarly used to evaluate the reproducibility of policies obtained via imitation.

% Behavioural Cloning (BC)~\cite{bain1995framework, giusti2015machine, bojarski2016end} is a popular and strong baseline used due to its simple mean-square error (MSE) loss over the dataset of transitions provided by the expert/provided trajectories. 
% However, it requires a large dataset that spans large parts of the possible state space to be effective as the policy can diverge due to errors and sources of noise in the environment.
% DAgger~\cite{ross2011reduction} is an approach which introduces an iterative data collection phase to correct for when the policy diverges using queries to an expert. 
%, although a BC or imitation approach could be used in a second phase once a policy is learnt.

\section{Conclusion}

In summary, our work highlights policy reproducibility, an important but commonly overlooked issue in evaluation of RL policies in uncertain environments.
We define policy reproducibility as the dispersion of the distribution of returns of a policy and propose two robust statistical measures to quantify this: Mean Absolute Deviation (MAD) and the Interquartile Range (IQR). 
To account for reproducibility when comparing and evaluating policies in RL, we advocate for the use of the Lower Confidence Bound (LCB) of the return distribution as an alternative to the commonly used mean return.
The LCB allows practitioners to set their preference for reproducibility with a single parameter $\alpha$ and is more general, as the mean return can also be recovered if $\alpha=0$.
Finally, by extensively evaluating popular RL algorithms, we experimentally show that there exists a performance-reproducibility trade-off which can be covered and effectively evaluated when using the proposed LCB metric.

%% file: figures/mr_wr_table.tex
\begin{table} 
\small
\centering
\begin{tabular}{c|c|cccc}

Env                         & Noise         & ES    & R-ES  & SAC       &     TD3 \\
\midrule
\multirow{5}{1.0cm}{Ant}    & Action        & 0.87  & 0.55  & 119.22    &  107.94 \\
                            & Init-State    & 0.31  & 0.11  & 1.70      &    0.74 \\
                            & Obs           & 0.72  & 0.15  & 56.34     &  143.77 \\
                            & Param         & 0.02  & 0.00  & 1.74      &    1.20 \\
                            & Reward        & 0.34  & 0.10  & 78.34     &   79.89 \\
\midrule
\multirow{5}{1.0cm}{Half\\Cheetah}  & Action        &  182.33 & 134.01  &  209.95 &  139.37 \\
                                    & Init-State    &  154.06 & 100.33  &  150.87 &  100.32 \\
                                    & Obs           &  192.72 & 126.61  &  279.67 &  229.85 \\
                                    & Param         &  200.30 & 134.96  &  125.78 &  174.45 \\
                                    & Reward        &  184.30 & 129.36  &  206.92 &  141.14 \\
\bottomrule
\end{tabular}
\caption{MAD scores of all algorithms on all environments and noise configurations. We report the IQM across 10 seeds. The lower the MAD score, the more reproducible the policy.}
\label{tab:mad_results}
\end{table}

%% file: figures/lcb_table.tex
\begin{table*} 
\footnotesize
\centering
\begin{tabular}{l|l|rrrr|rrrr}
\toprule
\multirow{2}{*}{Env}  & \multirow{2}{*}{Noise} & \multicolumn{4}{c|}{$\alpha= 0.0$} & \multicolumn{4}{c}{$\alpha=2.0\times10^3$} \\
            &  &      SAC &      TD3 &       ES & R-ES &       SAC &       TD3 &        ES & R-ES \\
\midrule
\multirow{5}{*}{Ant} & Action &  \textbf{6751.87} &  5002.73 &  4349.05 &  4789.71 &  -231613.46 &   -211449.83 &   2578.14 &  \textbf{3663.98} \\
                    & Init-State &  \textbf{7269.30} &  4796.49 &  5306.23 & 5155.73 &   3646.34 &   2393.72 &   4627.16 &  \textbf{4865.26} \\
                    & Obs &  -184.10 &   158.18 &  4593.58 &  \textbf{4644.19} &  -112922.20 &  -287392.25 &   2501.99 &  \textbf{4093.65} \\
                    & Param &  -536.90 &  -429.24 &   -17.73 & \textbf{6.11} &  -4035.81 & -2901.43 &   -50.10 &  \textbf{6.11} \\
                    & Reward &  \textbf{8115.54} &  5917.15 &  5234.34 & 5345.27 & -148633.70 &   -153633.47 &   4152.09 & \textbf{5042.58} \\
\midrule
 & & \multicolumn{4}{c|}{$\alpha= 0.0$} & \multicolumn{4}{c}{$\alpha=2.0\times10^2$} \\
 \midrule
\multirow{5}{*}{HalfCheetah} & Action &  \textbf{4700.75} &  3462.12 &    2212.68 & 2818.90 &  -36949.3 & -24171.7 &  -33946.0 &   \textbf{-23400.6} \\
                            & Init-State &  \textbf{5656.20} &  3399.78 &  2456.96 &  2450.23 &   -24634.6 &  \textbf{16278.7} & -28458.4 &  -17322.4 \\
                            & Obs &  2505.37 &  2858.40 &  2782.10 & \textbf{2947.29} &  -53471.0 &  -42928.2 &  -35405.4 &  \textbf{-21383.4} \\
                            & Param &  -208.22 &   315.36 &  1562.45 & \textbf{2827.84} &  -25366.7 &  -34592.7 &  -37925.7 & \textbf{-24100.1} \\
                            & Reward &  \textbf{5622.17} &  3217.24 &  2348.76 & 3200.21 &   -35460.5 & -24493.9 &   -34150.5 & \textbf{-22274.4} \\
\bottomrule
\end{tabular}

\caption{
LCB of all algorithms across environments and noise configurations for different values of $\alpha$. We report the IQM across 10 seeds. The higher the LCB, the better.
}
\label{tab:lcb_results}
\end{table*}

%% file: appendix.tex
\section{Experimental Details} \label{app:hyperparams}
All environments are run with an episode length of 1000 timesteps.
We use policies with two hidden layers of size 64 for all algorithms.
ES variants are run with mirror sampling and rank normalization as done in ~\citet{salimans2017evolution}. 
Reproducibility-ES optimises a weighted sum of performance and reproducibility, with equal weight for the two objectives (i.e. weight $50\%$).
Hyperparameters for all algorithms considered are provided in the following tables.

\input{hp_table}

\section{LCB - Supplementary Results} \label{app:lcb}

Results in Section 4.2 display the LCB using expected return for performance $P_{\pi}$ and MAD for reproducibility $\sigma_{\pi}$. 
In this section we use alternative metrics for $P_{\pi}$ and $\sigma_{\pi}$ as described in Section~\ref{subsec:quant_rep}.

\subsection{Using median to quantify performance}

\begin{figure}[h!]
\setlength{\abovecaptionskip}{-1pt} 
\setlength{\belowcaptionskip}{-10pt} 
\centering
    \includegraphics[width = 0.95\linewidth]{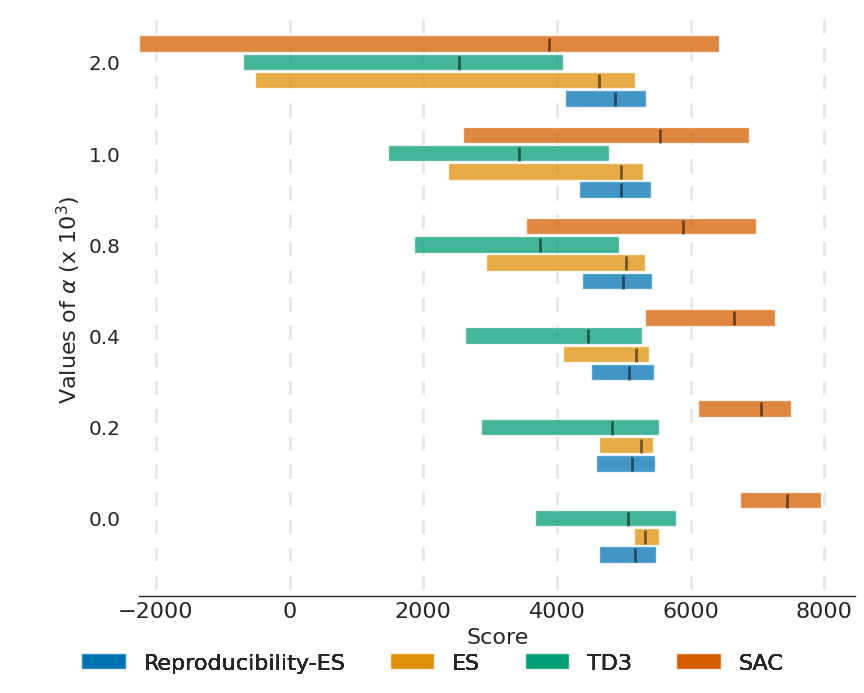}
    \caption{
        LCB comparison using \textbf{median} for performance $P_{\pi}$ and \textbf{MAD} for reproducibility $\sigma_{\pi}$, on the Ant environment with Init-State noise. 
        y-axis corresponds to varying values of $\alpha$, trade-off between performance and reproducibility. We report the IQM and the CIs across 10 seeds. The higher the LCB score the better.
    }
\label{fig:lcb_median}
\end{figure}

As highlighted in Section 3.2 the median is more robust to outliers than the expectation. 
Thus, we display the LCB using the median of the return to estimate the performance $P_{\pi}$ in Figure~\ref{fig:lcb_median} and Table~\ref{tab:lcb_median_results}. 
The general trend obtained with the median are similar to the ones obtained with the expectation in Figure 6 and Table 2, where the LCB scores of more reproducible policies, such as those from R-ES are less affected as alpha increases than those from SAC. 
%however the algorithms ordering is slightly modified for the HalfCheetah task. 

\subsection{Using STD or IQR to quantify reproducibility}

\begin{figure}[h!]
\setlength{\abovecaptionskip}{-1pt} 
\setlength{\belowcaptionskip}{-10pt} 
\centering
    \includegraphics[width = 0.95\linewidth]{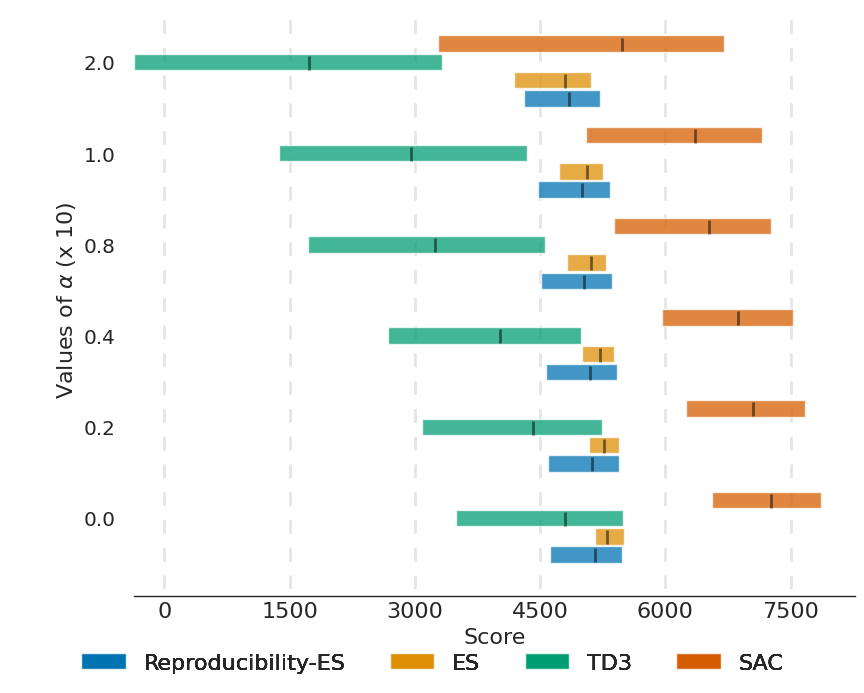}
    \caption{
        LCB comparison using \textbf{expected return} for performance $P_{\pi}$ and \textbf{IQR} for reproducibility $\sigma_{\pi}$, on the Ant with Init-State noise. 
        y-axis corresponds to varying values of $\alpha$, trade-off between performance and reproducibility. We report the IQM and the CIs across 10 seeds. The higher the LCB score the better.
    }
\label{fig:lcb_iqr}
\end{figure}

\begin{figure}[h!]
\setlength{\abovecaptionskip}{-1pt} 
\setlength{\belowcaptionskip}{-10pt} 
\centering
    \includegraphics[width = 0.95\linewidth]{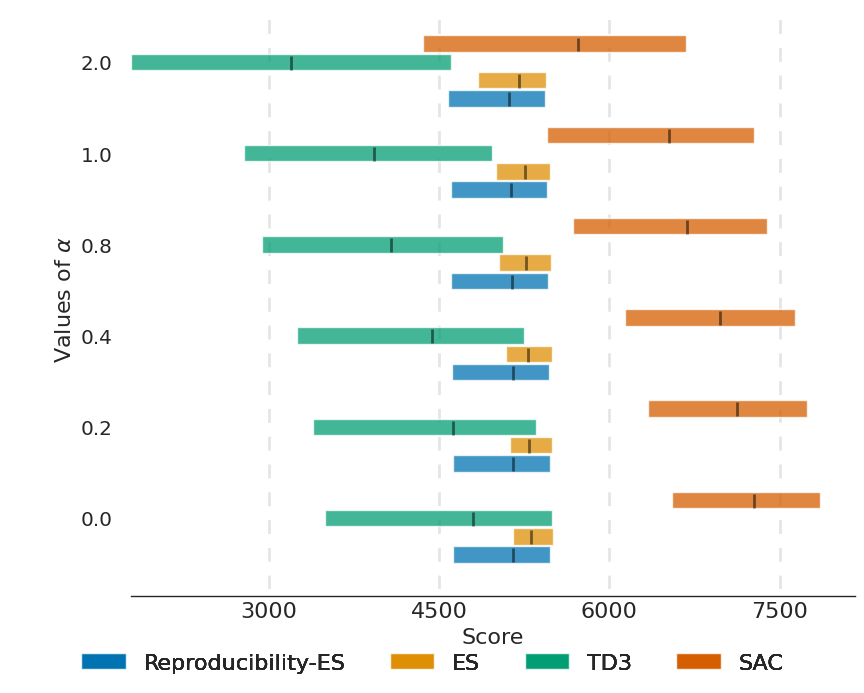}
    \caption{
        LCB comparison using \textbf{expected return} for performance $P_{\pi}$ and \textbf{standard deviation} for reproducibility $\sigma_{\pi}$, on the Ant with Init-State noise. 
        y-axis corresponds to varying values of $\alpha$, trade-off between performance and reproducibility. We report the IQM and the CIs across 10 seeds. The higher the LCB score the better.
    }
\label{fig:lcb_std}
\end{figure}

As mentioned in Section 3.2, the reproducibility $\sigma_{\pi}$ can also be quantified using the standard deviation or the IQR, another estimator robust to outliers. 
We give the LCB results using these alternatives in Figure~\ref{fig:lcb_iqr} and Table~\ref{tab:lcb_iqr_results} for IQR and in Figure~\ref{fig:lcb_std} and Table~\ref{tab:lcb_std_results} for standard deviation respectively. The results also demonstrate similar trends to the ones obtained with MAD estimation but at a different scale, indicating that in all cases of dispersion measures, our observations on the performance-reproducibility trade-off of the policies remain the same.

\input{figures/appendix_lcb_median_table}

\section{Reproducibility - Supplementary Results} \label{app:wr}

This section provide additional reproducibility quantification results to complement the results from Section 4.2.

\subsection{Return MAD}

\begin{figure}[h!]
\setlength{\abovecaptionskip}{-1pt} 
\setlength{\belowcaptionskip}{-10pt} 
\centering
    \includegraphics[width = 0.95\linewidth]{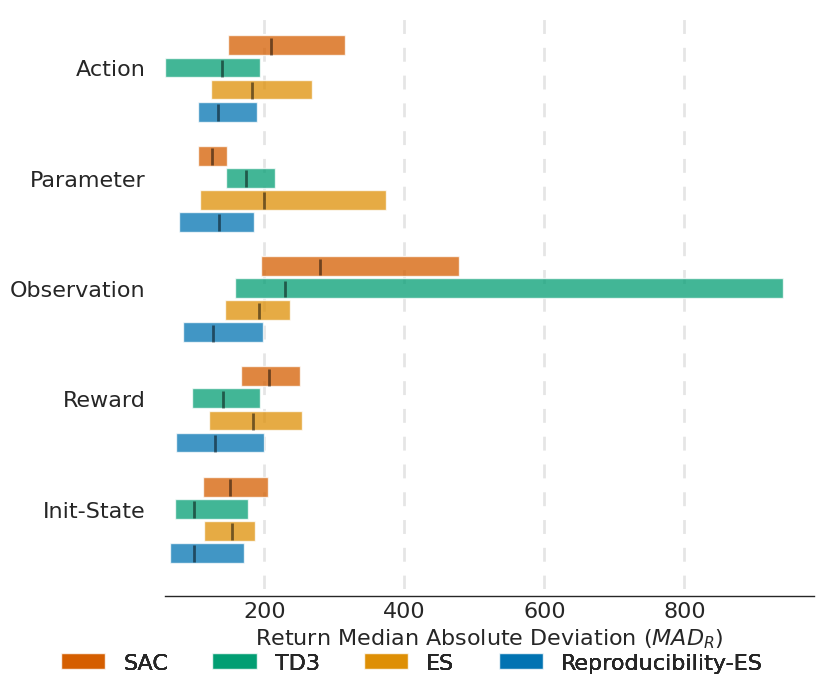}
    \caption{
        MAD scores of final policies on the HalfCheetah environment. y-axis the type of uncertainty present in the environment.  
        We report the IQM and the CIs across 10 seeds.
        The lower the MAD score the more reproducible the policy.
    }
\label{fig:hc_mad}
\end{figure}

We first provide in Figure~\ref{fig:hc_mad} the MAD results for the HalfCheetah environment, as Figure 5 only displays the results for the Ant environment. 
The numerical values for this metric can be found in Table 1.

\subsection{Return IQR}

\begin{figure}[h!]
\setlength{\abovecaptionskip}{-1pt} 
\setlength{\belowcaptionskip}{-10pt} 
\centering
    \includegraphics[width = 0.95\linewidth]{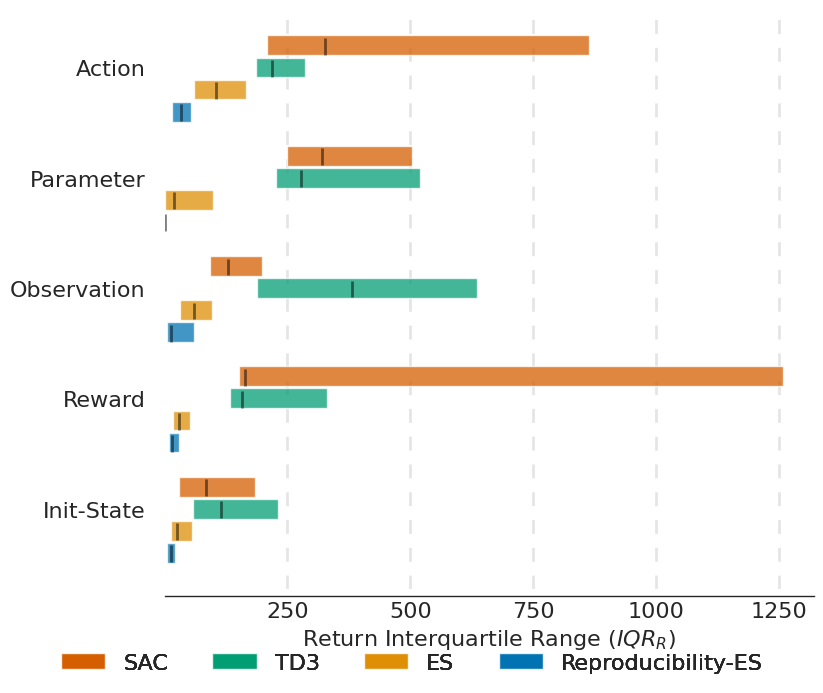}
    \caption{
        IQR scores of final policies in the Ant environment. y-axis the type of uncertainty present in the environment.  
        We report the IQM and the CIs across 10 seeds.
        The lower the IQR score the more reproducible the policy.
    }
\label{fig:wr_ant}
\end{figure}

\begin{figure}[h!]
\setlength{\abovecaptionskip}{-1pt} 
\setlength{\belowcaptionskip}{-10pt} 
\centering
    \includegraphics[width = 0.95\linewidth]{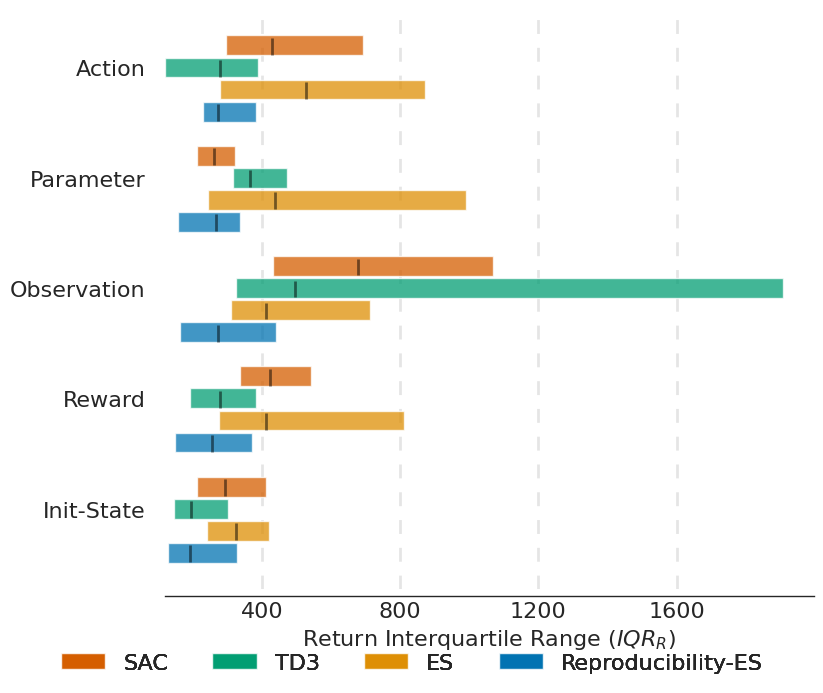}
    \caption{
        IQR scores of final policies in the HalfCheetah environment. y-axis the type of uncertainty present in the environment.  
        We report the IQM and the CIs across 10 seeds.
        The lower the IQR score the more reproducible the policy.
    }
\label{fig:wr_hc}
\end{figure}

As explained in Section 3.2, IQR is another measure of statistical dispersion of the distribution. It provides a better indication of the worst-case as it provides a wider representation of the distribution.
We provide in Figure~\ref{fig:wr_ant} and Figure~\ref{fig:wr_hc} the IQR results for Ant and HalfCheetah respectively.

For both tasks, the values of IQR are significantly higher than the value of MAD, which is expected as they represent the worse case. 
Interestingly, for the two environments, using the IQR instead of MAD leads to bigger confidence intervals, meaning larger range of values taken by different seeds. 
This is particularly visible for TD3 and SAC but can also be observed for ES. Additionally, the distance between TD3 and SAC and ES and Reproducibility-ES is reduced when using this new metric. However, ES and Reproducibility-ES remain the algorithms that result in the more reproducible policies even when considering the worst-case in the form of IQR.

\section{Behaviour Reproducibility - Supplementary Results} \label{app:bd}

\subsection{Behaviour Descriptor Representation}

\begin{figure}[h!]
\setlength{\abovecaptionskip}{-1pt} 
\setlength{\belowcaptionskip}{-10pt} 
\centering
    \includegraphics[width = 0.95\linewidth]{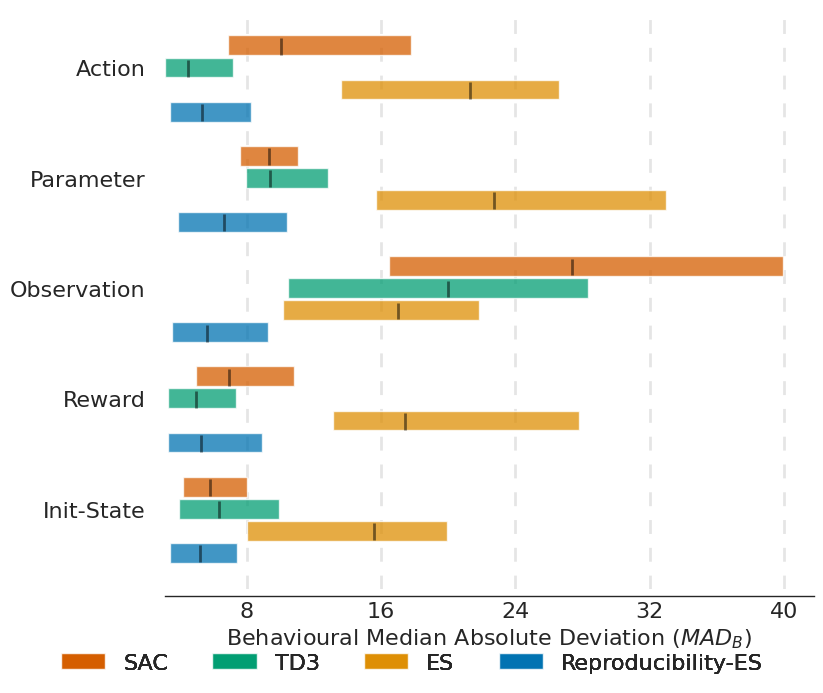}
    \caption{
    \textbf{Behaviour MAD} scores using the \textbf{descriptor} as behaviour representation of final policies in the HalfCheetah environment. 
    y-axis the type of uncertainty.  
    The lower the MAD score the more reproducible the behaviour.
    }
\label{fig:mr_bd_hc}
\end{figure}

\begin{figure}[h!]
\setlength{\abovecaptionskip}{-1pt} 
\setlength{\belowcaptionskip}{-10pt} 
\centering
    \includegraphics[width = 0.95\linewidth]{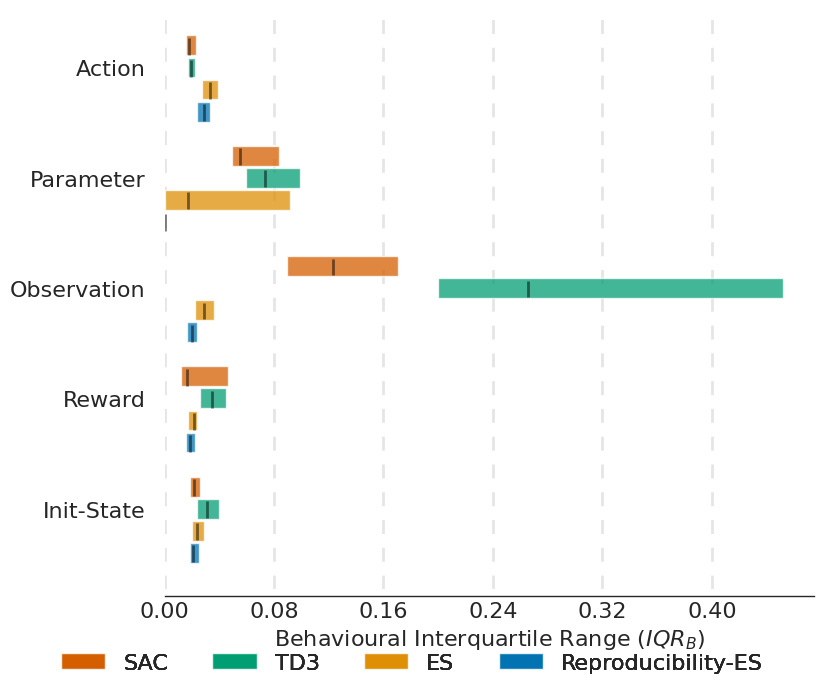}
    \caption{
    \textbf{Behaviour IQR} scores using the \textbf{descriptor} as behaviour representation of final policies in the Ant environment. 
    y-axis the type of uncertainty. 
    The lower the IQR score the more reproducible the behaviour.
    }
\label{fig:wr_bd_ant}
\end{figure}

\begin{figure}[h!]
\centering
    \includegraphics[width = 0.95\linewidth]{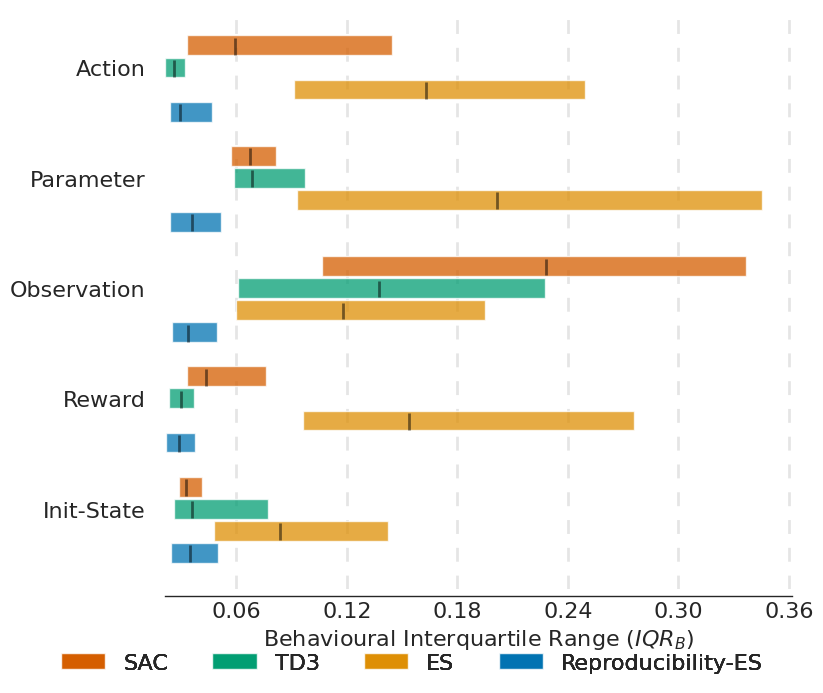}
    \caption{
    \textbf{Behaviour IQR} scores using the \textbf{descriptor} as behaviour representation of final policies in the HalfCheetah environment. 
    y-axis the type of uncertainty.  
    The lower the IQR score the more reproducible the behaviour.
    }
\label{fig:wr_bd_hc}
\end{figure}

As explained in Section~\ref{sec:behaviour}, the behavioural descriptor representation used in our experiments are taken from Novelty Search and Quality Diversity literature.
The average feet contact over the episode is used to characterise the gait of the robot~\cite{cully2015robots}. Hence, the descriptor is 4-dimensional for the Ant and 2-dimensional for the HalfCheetah.
We compute the MAD and IQR over the matrix of L2 distances in these spaces. 
We give the results for MAD in HalfCheetah environment in Figure~\ref{fig:mr_bd_hc} as well as the IQR for both environments in Figure~\ref{fig:wr_bd_ant} and Figure~\ref{fig:wr_bd_hc}.

While TD3, SAC and Reproducibility-ES seem to maintain their reproducibility results and trends in the HalfCheetah environment, ES solutions appears significantly less reproducible than in Ant. This is consistent with the results obtained for the return reproducibility. 
The IQR results yield the same conclusions as the MAD with larger confidence intervals, as for the return reproducibility.

\subsection{Results for State Marginal Representation}

As mentioned in Section 5, the state marginal can alternatively be used as behaviour representation. 
To compute reproducibility metrics using this representation, we consider all the states encounter by a policy during a full episode and concatenate them, giving a $(episode\_length * num\_state\_dimension)$ vector for each policy. 
We then compute the matrix of L2 distances of all pairs of these vectors, and use the MAD and IQR of these distances as reproduciblity metrics.

\begin{figure}[h!]
\centering
    \includegraphics[width = 0.95\linewidth]{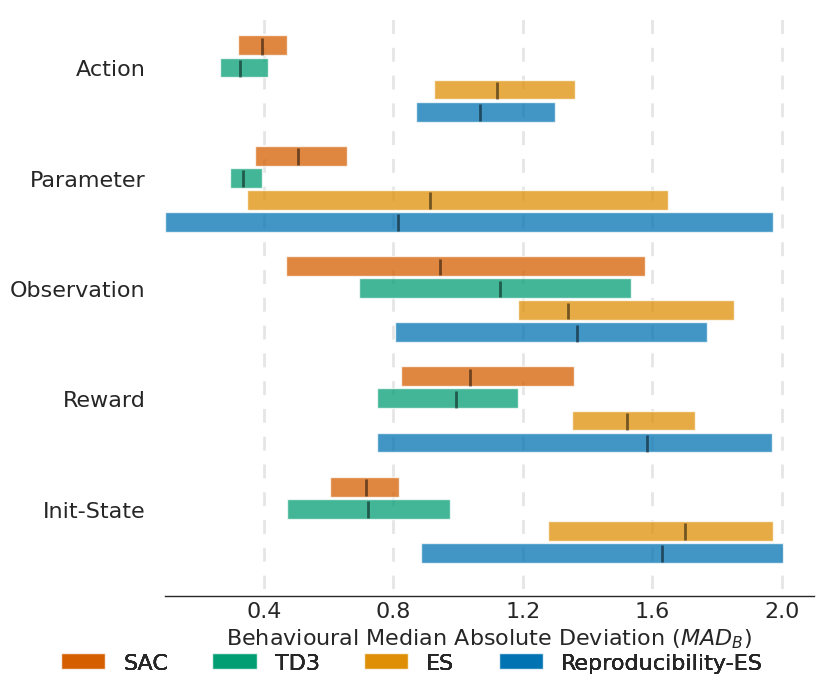}
    \caption{
        \textbf{Behaviour MAD} scores using \textbf{state-marginal} as behaviour representation of final policies in the Ant environment. 
        y-axis the type of uncertainty.
        The lower the MAD score the more reproducible the behaviour.
        }
\label{fig:results_stateMBR_ant}
\end{figure}

\begin{figure}[h!]
\centering
    \includegraphics[width = 0.95\linewidth]{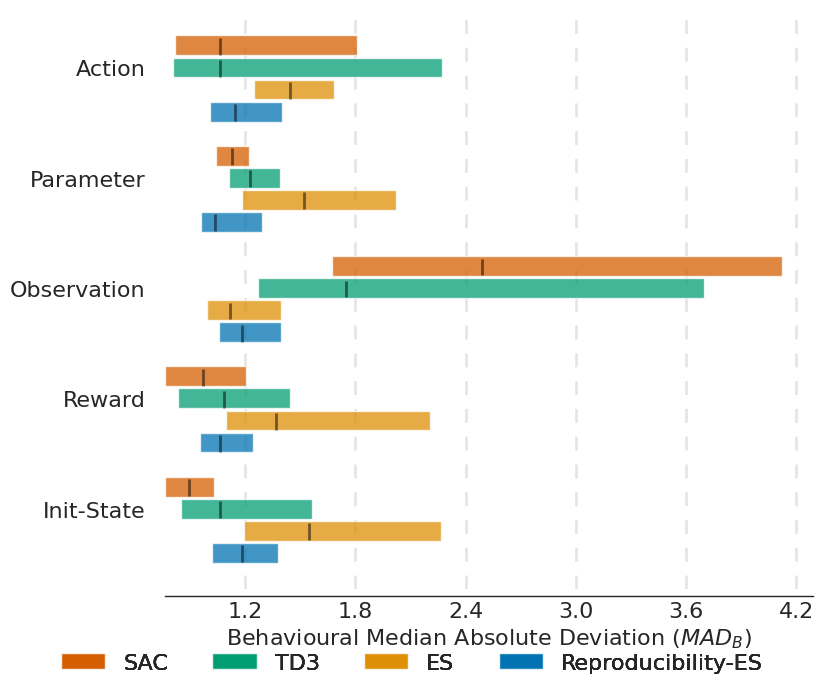}
    \caption{
        \textbf{Behaviour MAD} scores using \textbf{state-marginal} as behaviour representation of final policies in the HalfCheetah environment. 
        y-axis the type of uncertainty.  
        The lower the MAD score the more reproducible the behaviour.
    }
\label{fig:results_stateMBR_hc}
\end{figure}

\begin{figure}[h!]
\centering
    \includegraphics[width = 0.95\linewidth]{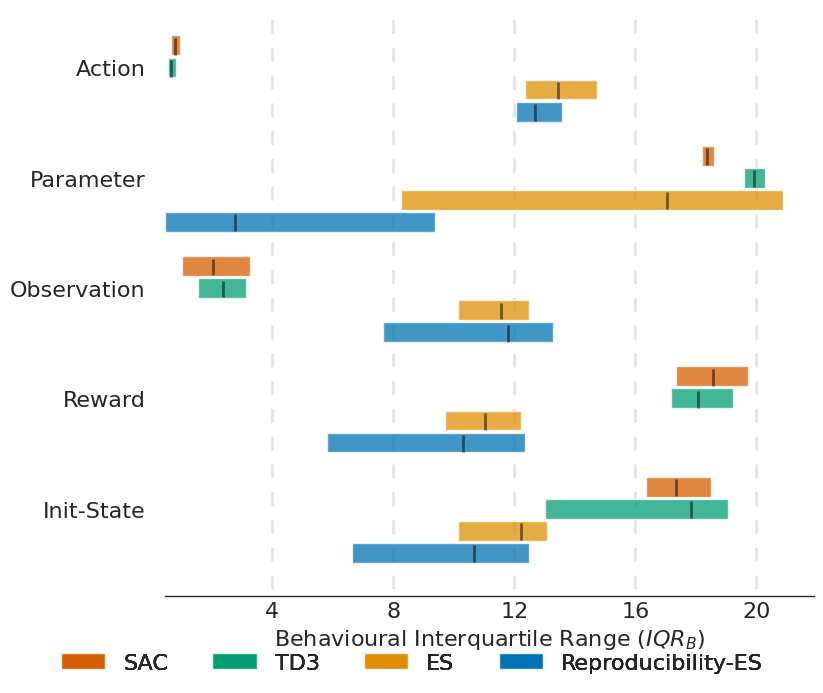}
    \caption{
        \textbf{Behaviour IQR} scores using \textbf{state-marginal} as behaviour representation of final policies in the Ant environment. 
        y-axis the type of uncertainty.  
        The lower the IQR score the more reproducible the behaviour.
    }
\label{fig:results_stateWBR_ant}
\end{figure}

\begin{figure}[h!]
\centering
    \includegraphics[width = 0.95\linewidth]{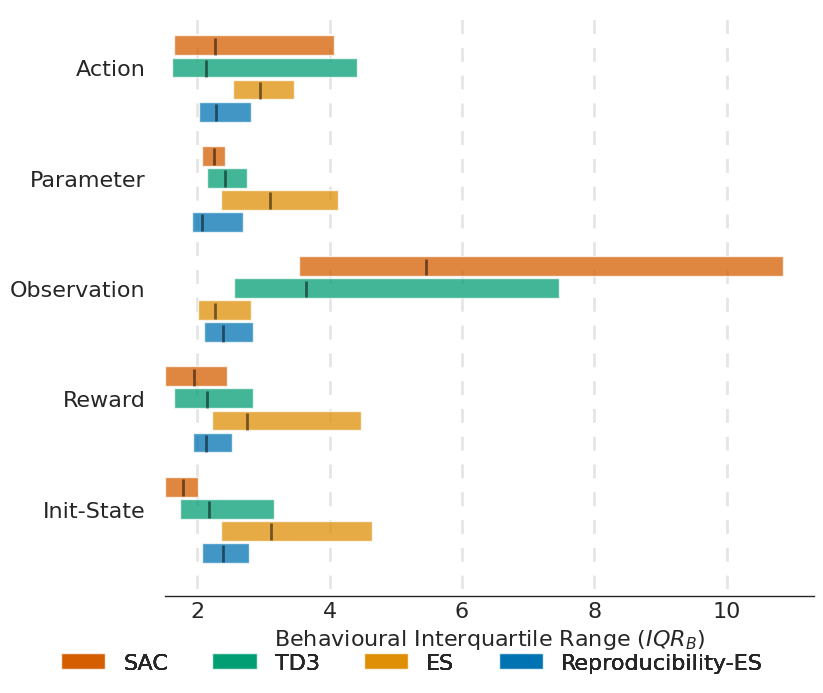}
    \caption{
        \textbf{Behaviour IQR} scores using \textbf{state-marginal} as behaviour representation of final policies in the HalfCheetah environment. 
        y-axis the type of uncertainty.  
        The lower the IQR score the more reproducible the behaviour.
    }
\label{fig:results_stateWBR_hc}
\end{figure}

We give the results for MAD and IQR for Ant in Figure~\ref{fig:results_stateMBR_ant} and Figure~\ref{fig:results_stateWBR_ant} respectively, and for HalfCheetah in Figure~\ref{fig:results_stateMBR_hc} and Figure~\ref{fig:results_stateWBR_hc} respectively.
With this representation, the two ES approaches lead to less reproducible policies than with the return and the behaviour descriptor representations. 
Additionally, this representation seems to widen the gap between MAD and IQR quantification. 
As state marginal is higher-dimensional, the distances are likely to be an order of magnitude larger than with the behaviour descriptor representation, leading to greater distances between worse and best case, as quantified by the IQR.
While providing an indication of the behaviour, the higher-dimensionality from the state-marginal also makes this distance estimation less representative. Literature has generally use lower-dimensional state-representations when computing the state-marginal distribution.

\section{Multi-objective Comparison} \label{app:pareto}

\begin{figure}[h!]
\centering
    \includegraphics[width = 0.95\linewidth]{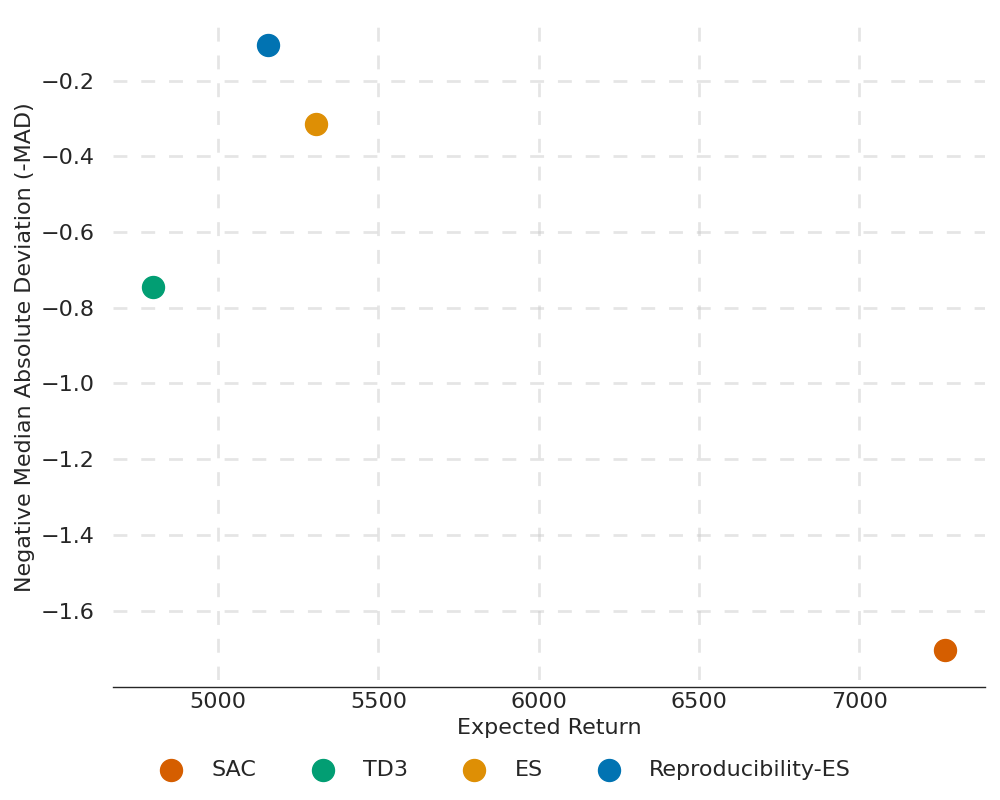}
    \caption{
        Pareto plot of performance and reproducibility in the Ant environment with Init-State noise. x-axis is the \textbf{expected return}, quantifying performance and y-axis is \textbf{-MAD}, quantifying reproducibility. Best approaches are the ones located close to the top-right corner.
    }
\label{fig:mo_ant_pareto}
\end{figure}

\begin{figure}[h!]
\centering
    \includegraphics[width = 0.95\linewidth]{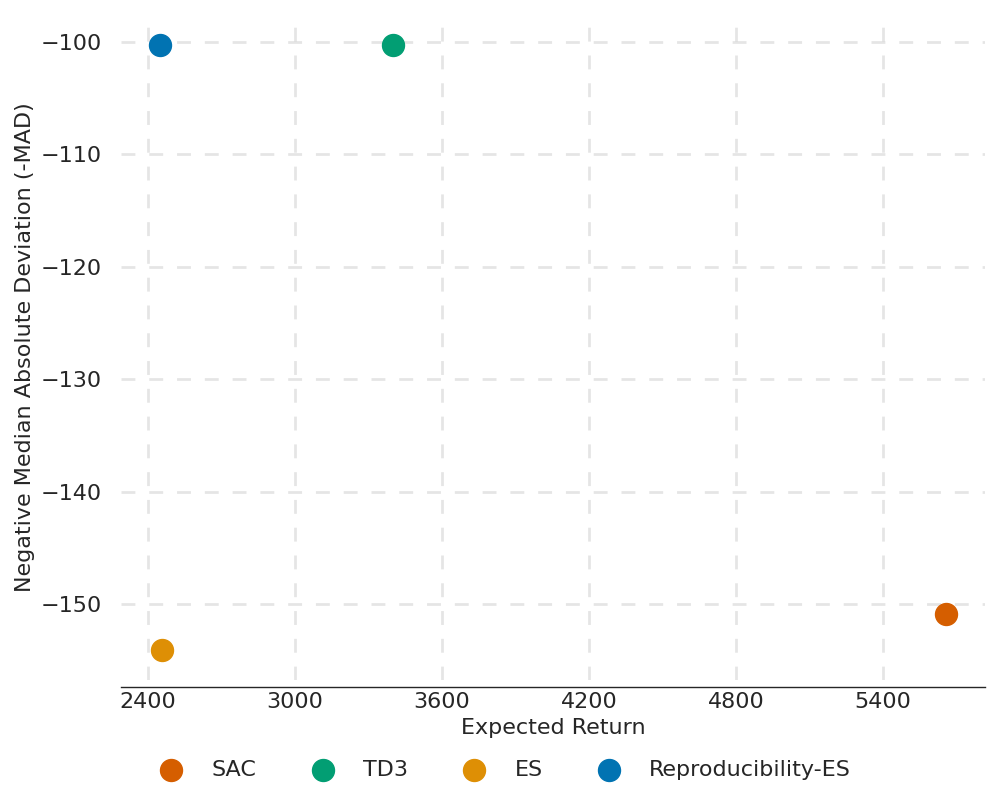}
    \caption{
        Pareto plot of performance and reproducibility in the HalfCheetah environment with Init-State noise. x-axis is the \textbf{expected return}, quantifying performance and y-axis is \textbf{-MAD}, quantifying reproducibility. Best approaches are the ones located close to the top-right corner.
    }
\label{fig:mo_hc_pareto}
\end{figure}

This section displays the trade-off between the performance $P_\pi$ and the reproducibility $\sigma_\pi$ of a policy from the lense of multi-objective optimization.
To do this, we plot in Figure~\ref{fig:mo_ant_pareto} and in Figure~\ref{fig:mo_hc_pareto} the pareto plots where $P_\pi$ is measured by the expected return and $\sigma_\pi$ is measured by the return MAD. 
On those plots, an approach dominates another if it performs better accordingly to at least one of the objectives. 
Pareto-dominant approaches are approaches that are not dominated by any other. The Pareto-front is the set of all Pareto-dominant solutions.
SAC and R-ES are both Pareto-dominant in Ant and HalfCheetah, each at one extreme of the Pareto front, thanks to their high performance and high reproducibility respectively.
This recovers the results obtained with the LCB in Section 5, where SAC is dominant when $\alpha$ set the focus fully on performance, and R-ES is dominant when $\alpha$ set the focus fully on reproducibility.

%% file: hp_table.tex
\begin{table}[htb]
\centering
\begin{tabular}{l|cc}
\toprule
\textsc{Parameter} & \textsc{Ant} & \textsc{Halfcheetah} \\
\midrule
\addlinespace
Critic hidden layer size & [256, 256] & [256, 256] \\
Replay buffer size & 1000000 & 1000000  \\
Critic learning rate & 0.0003 &  0.0003  \\ 
Actor learning rate &  0.0003 & 0.0003  \\ 
Noise clip &  0.5 & 0.5 \\
Policy noise &  0.2 & 0.2 \\
Discount $\gamma$ &  0.99 & 0.99  \\
Reward scaling &  1.0 & 1.0 \\
Batch size &  256 & 256  \\
Soft tau update &  0.005 & 0.005 \\
\bottomrule
\end{tabular}
\caption{
    Hyperparameters of TD3.
}
\label{tab:hyperparams_td3}
\end{table}

\begin{table}[htb]
\centering
\begin{tabular}{l|cc}
\toprule
\textsc{Parameter} & \textsc{Ant} & \textsc{Halfcheetah} \\
\midrule
\addlinespace
Critic hidden layer size & [256, 256] & [256, 256] \\
Replay buffer size & 1000000 & 1000000  \\
Critic learning rate & 0.0003  &  0.0003  \\ 
Actor learning rate &  0.0003  & 0.0003  \\ 
$\tau$ & 0.005 & 0.005 \\
Discount $\gamma$ &  0.99 & 0.99  \\
Reward scaling &  1.0 & 1.0 \\
Batch size &  256 & 256  \\
\bottomrule
\end{tabular}
\caption{
    Hyperparameters of SAC.
}
\label{tab:hyperparams_sac}
\end{table}

\begin{table}[htb]
\centering
\begin{tabular}{l|cc}
\toprule
\textsc{Parameter}& \textsc{Ant} & \textsc{Halfcheetah} \\
\midrule
\addlinespace
Population size & 512 & 512 \\
Learning rate &  0.005 & 0.001  \\ 
Perturbation $\sigma$ &  0.02 & 0.01 \\
L2 coefficient &  0.0 & 0.0 \\

\bottomrule
\end{tabular}
\caption{
    Hyperparameters of ES and Reproducibility-ES.
}
\label{tab:hyperparams_es}
\end{table}

%% file: figures/appendix_lcb_median_table.tex
\begin{table*} 
\footnotesize
\centering
\begin{tabular}{l|l|rrrr|rrrr}
\toprule
\multirow{2}{*}{Env}  & \multirow{2}{*}{Noise} & \multicolumn{4}{c|}{$\alpha= 0.0$} & \multicolumn{4}{c}{$\alpha=2.0\times10^3$} \\
            &  &      SAC &      TD3 &       ES & R-ES &       SAC &       TD3 &        ES & R-ES \\
\midrule
\multirow{5}{*}{Ant} & Action &  \textbf{7360.33} &  5149.47 &  4386.08 &  4785.91 &  -230983.67 &   -211329.97 &   2664.24 &  \textbf{3653.98} \\
                    & Init-State &  \textbf{7436.86} &  5057.88 &  5312.11 & 5161.15 &   3877.37 &   2529.72 &   4635.63 &  \textbf{4868.98} \\
                    & Obs &  -202.17 &   220.66 &  4597.87 &  \textbf{4629.19} &  -112939.07 &  -287292.89 &   2518.08 &  \textbf{4078.65} \\
                    & Param &  -509.63 &  -298.95 &   -8.46 & \textbf{6.11} &  -3983.13 & -2792.61 &   -40.82 &  \textbf{6.11} \\
                    & Reward &  \textbf{8196.79} &  5937.11 &  5223.86 & 5341.68 & -148566.25 &   -153614.93 &   4158.34 & \textbf{5043.34} \\
\midrule
 & & \multicolumn{4}{c|}{$\alpha= 0.0$} & \multicolumn{4}{c}{$\alpha=2.0\times10^2$} \\
 \midrule
\multirow{5}{*}{HalfCheetah} & Action &  \textbf{4700.75} &  3462.12 &    2212.68 & 2818.90 &  -36949.3 & -24171.7 &  -33946.0 &   \textbf{-23400.6} \\
                            & Init-State &  \textbf{5800.36} &  3537.65 &  2669.87 &  2482.52 & -295965.34 &  \textbf{-196825.05} & -305615.85 &  -197912.01 \\
                            & Obs & 2719.77 & \textbf{3103.79} & 3094.15 & 2993.59 &  -556693.14 & -456379.18 & -382018.16 &  \textbf{-249007.95} \\
                            & Param & -244.44 & 275.96 & 1819.47 & \textbf{2917.69} &  \textbf{-251791.34} & -348607.86 & -398284 & -266958.19 \\
                            & Reward &  \textbf{5760.14} &  3283.46 &  2721.6 & 3291.01 &   -408010.79 & -278541.65 & -365753.28 & \textbf{-255068.67} \\
\bottomrule
\end{tabular}

\caption{
LCB results using \textbf{median} for performance $P_{\pi}$ and \textbf{MAD} for reproducibility $\sigma_\pi$. We report the IQM across 10 seeds. The higher the LCB, the better.
}
\label{tab:lcb_median_results}

\end{table*}

\begin{table*} 
\footnotesize
\centering
\begin{tabular}{l|l|rrrr|rrrr}
\toprule
\multirow{2}{*}{Env}  & \multirow{2}{*}{Noise} & \multicolumn{4}{c|}{$\alpha= 0.0$} & \multicolumn{4}{c}{$\alpha=2.0\times10^1$} \\
            &  &      SAC &      TD3 &       ES & R-ES &       SAC &       TD3 &        ES & R-ES \\
\midrule
\multirow{5}{*}{Ant} & Action &  \textbf{6751.87} &  5002.73 &  4349.05 &  4789.71 &  267.35 &   447.17 &   2226.47 &  \textbf{4117.97} \\
                    & Init-State &  \textbf{7269.3} &  4796.49 &  5306.23 & 5155.73 &   \textbf{5481.64} &   1735.59 &   4803.18 &  4851.69 \\
                    & Obs &  -184.1 &   158.18 &  4593.58 &  \textbf{4644.19} &  -2740.59 &  -7605.6 &   2882.03 &  \textbf{3728.27} \\
                    & Param &  -536.9 &  -429.24 & -17.73 & \textbf{6.11} & -6956.74 & -5947.28 &  -420.59 &  \textbf{6.11} \\
                    & Reward &  \textbf{8115.54} &  5917.15 &  5234.34 & 5345.27 & 4732.08 &   2637.38 &   4374.88 & \textbf{5026.96} \\
\midrule
 & & \multicolumn{4}{c|}{$\alpha= 0.0$} & \multicolumn{4}{c}{$\alpha=2.0\times10^1$} \\
 \midrule
\multirow{5}{*}{HalfCheetah} & Action &  \textbf{4700.75} &  3462.12 &  2212.68 & 2818.9 & -3740.33 & \textbf{-1884.07} & -8096.1 & -2437.29 \\
                            & Init-State &  \textbf{5656.2} &  3399.78 &  2456.96 &  2450.23 & -780.01 &  \textbf{-537.53} & -4230.86 &  -1107.99 \\
                            & Obs & 2505.37 & \textbf{2858.4} & 2782.1 & 2947.29 & -11121.63 & -6880.1 & -5365.77 &  \textbf{-2066.48} \\
                            & Param & -208.22 & 315.36 & 1562.45 & \textbf{2827.84} &  -5461.27 & -7069.18 & -6814.3 & \textbf{-2391.74} \\
                            & Reward &  \textbf{5622.17} &  3217.24 &  2348.76 & 3200.21 &  -2633.95 & -1953.78 & -5919.28 & \textbf{-1512.99} \\
\bottomrule
\end{tabular}

\caption{
LCB results using \textbf{expected return} for performance $P_{\pi}$ and \textbf{IQR} for reproducibility $\sigma_\pi$. We report the IQM across 10 seeds. The higher the LCB, the better.
}
\label{tab:lcb_iqr_results}

\end{table*}

\begin{table*} 
\footnotesize
\centering
\begin{tabular}{l|l|rrrr|rrrr}
\toprule
\multirow{2}{*}{Env}  & \multirow{2}{*}{Noise} & \multicolumn{4}{c|}{$\alpha= 0.0$} & \multicolumn{4}{c}{$\alpha=2.0$} \\
            &  &      SAC &      TD3 &       ES & R-ES &       SAC &       TD3 &        ES & R-ES \\
\midrule
\multirow{5}{*}{Ant} & Action &  \textbf{6751.87} &  5002.73 &  4349.05 &  4789.71 & 3523.46 & 3520.11 & 3987.25 &  \textbf{4673.97} \\
                    & Init-State &  \textbf{7269.3} &  4796.49 &  5306.23 & 5155.73 &  5722.39 & 3199.28 &   5202.23 &  \textbf{5115.46} \\
                    & Obs & -184.1 &   158.18 &  4593.58 &  \textbf{4644.19} &  -499.39 & -316.85 &  4384.15 &  \textbf{4570.03} \\
                    & Param &  -536.9 &  -429.24 &  -17.73 & \textbf{6.11} &  -1006.84 & -1031.25 &  -69.06 &  \textbf{6.11} \\
                    & Reward &  \textbf{8115.54} & 5917.15 &  5234.34 & 5345.27 & 7286.9 &  5604.09 &  5140.04 & \textbf{5300.37} \\
\midrule
 & & \multicolumn{4}{c|}{$\alpha= 0.0$} & \multicolumn{4}{c}{$\alpha=2.0$} \\
 \midrule
\multirow{5}{*}{HalfCheetah} & Action &  \textbf{4700.75} &  3462.12 &  2212.68 & 2818.9 & \textbf{2661.66} & 2382.48 & 559.46 &   2054.48 \\
                            & Init-State &  \textbf{5656.2} &  3399.78 &  2456.96 &  2450.23 & \textbf{4238.76} &  2218.53 & 1048.04 &  2035.68 \\
                            & Obs & 2505.37 & 2858.4 & 2782.1 & \textbf{2947.29} & 834.48 & 810.48 & 1023.25 &  \textbf{2224.08} \\
                            & Param & -208.22 & 315.36 & 1562.45 & \textbf{2827.84} &  -652.38 & -330.38 & 178.39 & \textbf{1926.9} \\
                            & Reward &  \textbf{5622.17} &  3217.24 &  2348.76 & 3200.21 &  \textbf{4643.38} & 2375.69 & 783.2 & 2216.81 \\
\bottomrule
\end{tabular}

\caption{
LCB using \textbf{expected return} for performance $P_{\pi}$ and \textbf{standard deviation} for reproducibility $\sigma_{\pi}$. We report the IQM across 10 seeds. The higher the LCB, the better.
}
\label{tab:lcb_std_results}

\end{table*}

%% file: Formatting-Instructions-LaTeX-2024.bbl
\begin{thebibliography}{43}
\providecommand{\natexlab}[1]{#1}

\bibitem[{Agarwal et~al.(2021)Agarwal, Schwarzer, Castro, Courville, and
  Bellemare}]{agarwal2021deep}
Agarwal, R.; Schwarzer, M.; Castro, P.~S.; Courville, A.~C.; and Bellemare, M.
  2021.
\newblock Deep reinforcement learning at the edge of the statistical precipice.
\newblock \emph{Advances in neural information processing systems}, 34:
  29304--29320.

\bibitem[{Akkaya et~al.(2019)Akkaya, Andrychowicz, Chociej, Litwin, McGrew,
  Petron, Paino, Plappert, Powell, Ribas et~al.}]{akkaya2019solving}
Akkaya, I.; Andrychowicz, M.; Chociej, M.; Litwin, M.; McGrew, B.; Petron, A.;
  Paino, A.; Plappert, M.; Powell, G.; Ribas, R.; et~al. 2019.
\newblock Solving rubik's cube with a robot hand.
\newblock \emph{arXiv preprint arXiv:1910.07113}.

\bibitem[{Azar et~al.(2011)Azar, Munos, Ghavamzadeh, and
  Kappen}]{azar2011speedy}
Azar, M.~G.; Munos, R.; Ghavamzadeh, M.; and Kappen, H. 2011.
\newblock Speedy Q-learning.
\newblock In \emph{Advances in neural information processing systems}.

\bibitem[{Bellemare et~al.(2017)Bellemare, Dabney, and
  Munos}]{bellemare2017distributional}
Bellemare, M.~G.; Dabney, W.; and Munos, R. 2017.
\newblock A distributional perspective on reinforcement learning.
\newblock In \emph{International conference on machine learning}, 449--458.
  PMLR.

\bibitem[{Bellemare et~al.(2023)Bellemare, Dabney, and
  Rowland}]{bellemare2023distributional}
Bellemare, M.~G.; Dabney, W.; and Rowland, M. 2023.
\newblock \emph{Distributional reinforcement learning}.
\newblock MIT Press.

\bibitem[{Brunke et~al.(2022)Brunke, Greeff, Hall, Yuan, Zhou, Panerati, and
  Schoellig}]{brunke2022safe}
Brunke, L.; Greeff, M.; Hall, A.~W.; Yuan, Z.; Zhou, S.; Panerati, J.; and
  Schoellig, A.~P. 2022.
\newblock Safe learning in robotics: From learning-based control to safe
  reinforcement learning.
\newblock \emph{Annual Review of Control, Robotics, and Autonomous Systems}, 5:
  411--444.

\bibitem[{Cassandra(1998)}]{cassandra1998survey}
Cassandra, A.~R. 1998.
\newblock A survey of POMDP applications.
\newblock In \emph{Working notes of AAAI 1998 fall symposium on planning with
  partially observable Markov decision processes}, volume 1724.

\bibitem[{Chatzilygeroudis et~al.(2021)Chatzilygeroudis, Cully, Vassiliades,
  and Mouret}]{chatzilygeroudis2021quality}
Chatzilygeroudis, K.; Cully, A.; Vassiliades, V.; and Mouret, J.-B. 2021.
\newblock Quality-Diversity Optimization: a novel branch of stochastic
  optimization.
\newblock In \emph{Black Box Optimization, Machine Learning, and No-Free Lunch
  Theorems}, 109--135. Springer.

\bibitem[{Chen et~al.(2020)Chen and Li}]{chen2020overview}
Chen, S.; and Li, Y. 2020.
\newblock An overview of robust reinforcement learning.
\newblock In \emph{2020 IEEE International Conference on Networking, Sensing
  and Control (ICNSC)}, 1--6. IEEE.

\bibitem[{Coraluppi et~al.(1999)Coraluppi and Marcus}]{coraluppi1999risk}
Coraluppi, S.~P.; and Marcus, S.~I. 1999.
\newblock Risk-sensitive and minimax control of discrete-time, finite-state
  Markov decision processes.
\newblock \emph{Automatica}, 35(2): 301--309.

\bibitem[{Cully(2019)}]{cully2019autonomous}
Cully, A. 2019.
\newblock Autonomous skill discovery with quality-diversity and unsupervised
  descriptors.
\newblock In \emph{Proceedings of the Genetic and Evolutionary Computation
  Conference}, 81--89.

\bibitem[{Cully et~al.(2015)Cully, Clune, Tarapore, and
  Mouret}]{cully2015robots}
Cully, A.; Clune, J.; Tarapore, D.; and Mouret, J.-B. 2015.
\newblock Robots that can adapt like animals.
\newblock \emph{Nature}, 521(7553): 503--507.

\bibitem[{Dulac-Arnold et~al.(2020)Dulac-Arnold, Levine, Mankowitz, Li,
  Paduraru, Gowal, and Hester}]{dulac2020empirical}
Dulac-Arnold, G.; Levine, N.; Mankowitz, D.~J.; Li, J.; Paduraru, C.; Gowal,
  S.; and Hester, T. 2020.
\newblock An empirical investigation of the challenges of real-world
  reinforcement learning.
\newblock \emph{arXiv preprint arXiv:2003.11881}.

\bibitem[{Everitt et~al.(2017)Everitt, Krakovna, Orseau, Hutter, and
  Legg}]{everitt2017reinforcement}
Everitt, T.; Krakovna, V.; Orseau, L.; Hutter, M.; and Legg, S. 2017.
\newblock Reinforcement learning with a corrupted reward channel.
\newblock \emph{arXiv preprint arXiv:1705.08417}.

\bibitem[{Flageat et~al.(2023)Flageat and Cully}]{flageat2023uncertain}
Flageat, M.; and Cully, A. 2023.
\newblock Uncertain Quality-Diversity: Evaluation methodology and new methods
  for Quality-Diversity in Uncertain Domains.
\newblock \emph{IEEE Transactions on Evolutionary Computation}.

\bibitem[{Fox et~al.(2015)Fox, Pakman, and Tishby}]{fox2015taming}
Fox, R.; Pakman, A.; and Tishby, N. 2015.
\newblock Taming the noise in reinforcement learning via soft updates.
\newblock \emph{arXiv preprint arXiv:1512.08562}.

\bibitem[{Frazier(2018)}]{frazier2018tutorial}
Frazier, P.~I. 2018.
\newblock A tutorial on Bayesian optimization.
\newblock \emph{arXiv preprint arXiv:1807.02811}.

\bibitem[{Freeman et~al.(2021)Freeman, Frey, Raichuk, Girgin, Mordatch, and
  Bachem}]{freeman2021brax}
Freeman, C.~D.; Frey, E.; Raichuk, A.; Girgin, S.; Mordatch, I.; and Bachem, O.
  2021.
\newblock Brax--A Differentiable Physics Engine for Large Scale Rigid Body
  Simulation.
\newblock \emph{arXiv preprint arXiv:2106.13281}.

\bibitem[{Fujimoto et~al.(2018)Fujimoto, Hoof, and
  Meger}]{fujimoto2018addressing}
Fujimoto, S.; Hoof, H.; and Meger, D. 2018.
\newblock Addressing function approximation error in actor-critic methods.
\newblock In \emph{International conference on machine learning}, 1587--1596.
  PMLR.

\bibitem[{Garc{\i}a et~al.(2015)Garc{\i}a and
  Fern{\'a}ndez}]{garcia2015comprehensive}
Garc{\i}a, J.; and Fern{\'a}ndez, F. 2015.
\newblock A comprehensive survey on safe reinforcement learning.
\newblock \emph{Journal of Machine Learning Research}, 16(1): 1437--1480.

\bibitem[{Gu et~al.(2022)Gu, Yang, Du, Chen, Walter, Wang, Yang, and
  Knoll}]{gu2022review}
Gu, S.; Yang, L.; Du, Y.; Chen, G.; Walter, F.; Wang, J.; Yang, Y.; and Knoll,
  A. 2022.
\newblock A review of safe reinforcement learning: Methods, theory and
  applications.
\newblock \emph{arXiv preprint arXiv:2205.10330}.

\bibitem[{Haarnoja et~al.(2018)Haarnoja, Zhou, Abbeel, and
  Levine}]{haarnoja2018soft}
Haarnoja, T.; Zhou, A.; Abbeel, P.; and Levine, S. 2018.
\newblock Soft actor-critic: Off-policy maximum entropy deep reinforcement
  learning with a stochastic actor.
\newblock In \emph{International conference on machine learning}, 1861--1870.
  PMLR.

\bibitem[{Hansen(2006)}]{hansen2006cma}
Hansen, N. 2006.
\newblock The CMA evolution strategy: a comparing review.
\newblock \emph{Towards a new evolutionary computation: Advances in the
  estimation of distribution algorithms}, 75--102.

\bibitem[{Hasselt(2010)}]{hasselt2010double}
Hasselt, H. 2010.
\newblock Double Q-learning.
\newblock \emph{Advances in neural information processing systems}, 23.

\bibitem[{Heger(1994)}]{heger1994consideration}
Heger, M. 1994.
\newblock Consideration of risk in reinforcement learning.
\newblock In \emph{Machine Learning Proceedings 1994}, 105--111. Elsevier.

\bibitem[{Henderson et~al.(2018)Henderson, Islam, Bachman, Pineau, Precup, and
  Meger}]{henderson2018deep}
Henderson, P.; Islam, R.; Bachman, P.; Pineau, J.; Precup, D.; and Meger, D.
  2018.
\newblock Deep reinforcement learning that matters.
\newblock In \emph{Proceedings of the AAAI conference on artificial
  intelligence}, volume~32.

\bibitem[{Jin et~al.(2005)Jin and Branke}]{jin2005evolutionary}
Jin, Y.; and Branke, J. 2005.
\newblock Evolutionary optimization in uncertain environments-a survey.
\newblock \emph{IEEE Transactions on evolutionary computation}, 9(3): 303--317.

\bibitem[{Lehman et~al.(2018)Lehman, Chen, Clune, and Stanley}]{lehman2018more}
Lehman, J.; Chen, J.; Clune, J.; and Stanley, K.~O. 2018.
\newblock ES is more than just a traditional finite-difference approximator.
\newblock In \emph{Proceedings of the genetic and evolutionary computation
  conference}, 450--457.

\bibitem[{Lehman et~al.(2011)Lehman and Stanley}]{lehman2011abandoning}
Lehman, J.; and Stanley, K.~O. 2011.
\newblock Abandoning objectives: Evolution through the search for novelty
  alone.
\newblock \emph{Evolutionary computation}, 19(2): 189--223.

\bibitem[{Lynch et~al.(2020)Lynch, Khansari, Xiao, Kumar, Tompson, Levine, and
  Sermanet}]{lynch2020learning}
Lynch, C.; Khansari, M.; Xiao, T.; Kumar, V.; Tompson, J.; Levine, S.; and
  Sermanet, P. 2020.
\newblock Learning latent plans from play.
\newblock In \emph{Conference on robot learning}, 1113--1132. PMLR.

\bibitem[{Moos et~al.(2022)Moos, Hansel, Abdulsamad, Stark, Clever, and
  Peters}]{moos2022robust}
Moos, J.; Hansel, K.; Abdulsamad, H.; Stark, S.; Clever, D.; and Peters, J.
  2022.
\newblock Robust reinforcement learning: A review of foundations and recent
  advances.
\newblock \emph{Machine Learning and Knowledge Extraction}, 4(1): 276--315.

\bibitem[{Morimoto et~al.(2005)Morimoto and Doya}]{morimoto2005robust}
Morimoto, J.; and Doya, K. 2005.
\newblock Robust reinforcement learning.
\newblock \emph{Neural computation}, 17(2): 335--359.

\bibitem[{Parker-Holder et~al.(2020)Parker-Holder, Pacchiano, Choromanski, and
  Roberts}]{parker2020effective}
Parker-Holder, J.; Pacchiano, A.; Choromanski, K.~M.; and Roberts, S.~J. 2020.
\newblock Effective diversity in population based reinforcement learning.
\newblock \emph{Advances in Neural Information Processing Systems}, 33:
  18050--18062.

\bibitem[{Pugh et~al.(2016)Pugh, Soros, and Stanley}]{pugh2016quality}
Pugh, J.~K.; Soros, L.~B.; and Stanley, K.~O. 2016.
\newblock Quality diversity: A new frontier for evolutionary computation.
\newblock \emph{Frontiers in Robotics and AI}, 3: 40.

\bibitem[{Rakshit et~al.(2017)Rakshit, Konar, and Das}]{rakshit2017noisy}
Rakshit, P.; Konar, A.; and Das, S. 2017.
\newblock Noisy evolutionary optimization algorithms--a comprehensive survey.
\newblock \emph{Swarm and Evolutionary Computation}, 33: 18--45.

\bibitem[{Romoff et~al.(2018)Romoff, Henderson, Pich{\'e}, Francois-Lavet, and
  Pineau}]{romoff2018reward}
Romoff, J.; Henderson, P.; Pich{\'e}, A.; Francois-Lavet, V.; and Pineau, J.
  2018.
\newblock Reward estimation for variance reduction in deep reinforcement
  learning.
\newblock \emph{arXiv preprint arXiv:1805.03359}.

\bibitem[{Salimans et~al.(2017)Salimans, Ho, Chen, Sidor, and
  Sutskever}]{salimans2017evolution}
Salimans, T.; Ho, J.; Chen, X.; Sidor, S.; and Sutskever, I. 2017.
\newblock Evolution strategies as a scalable alternative to reinforcement
  learning.
\newblock \emph{arXiv preprint arXiv:1703.03864}.

\bibitem[{Sutton et~al.(2018)Sutton and Barto}]{sutton2018reinforcement}
Sutton, R.~S.; and Barto, A.~G. 2018.
\newblock \emph{Reinforcement learning: An introduction}.
\newblock MIT press.

\bibitem[{Tassa et~al.(2018)Tassa, Doron, Muldal, Erez, Li, Casas, Budden,
  Abdolmaleki, Merel, Lefrancq et~al.}]{tassa2018deepmind}
Tassa, Y.; Doron, Y.; Muldal, A.; Erez, T.; Li, Y.; Casas, D. d.~L.; Budden,
  D.; Abdolmaleki, A.; Merel, J.; Lefrancq, A.; et~al. 2018.
\newblock Deepmind control suite.
\newblock \emph{arXiv preprint arXiv:1801.00690}.

\bibitem[{Wang et~al.(2020)Wang, Liu, and Li}]{wang2020reinforcement}
Wang, J.; Liu, Y.; and Li, B. 2020.
\newblock Reinforcement learning with perturbed rewards.
\newblock In \emph{Proceedings of the AAAI conference on artificial
  intelligence}, volume~34, 6202--6209.

\bibitem[{Wierstra et~al.(2014)Wierstra, Schaul, Glasmachers, Sun, Peters, and
  Schmidhuber}]{wierstra2014natural}
Wierstra, D.; Schaul, T.; Glasmachers, T.; Sun, Y.; Peters, J.; and
  Schmidhuber, J. 2014.
\newblock Natural evolution strategies.
\newblock \emph{The Journal of Machine Learning Research}, 15(1): 949--980.

\bibitem[{Wilcox(2011)}]{wilcox2011introduction}
Wilcox, R.~R. 2011.
\newblock \emph{Introduction to robust estimation and hypothesis testing}.
\newblock Academic press.

\bibitem[{Xu et~al.(2006)Xu and Mannor}]{xu2006robustness}
Xu, H.; and Mannor, S. 2006.
\newblock The robustness-performance tradeoff in Markov decision processes.
\newblock \emph{Advances in Neural Information Processing Systems}, 19.

\end{thebibliography}
